\documentclass[onecolumn,3p]{elsarticle} %

\usepackage[pdfstartview={FitH}]{hyperref}
\usepackage[cmex10]{amsmath}

\usepackage{url}

\usepackage[usenames,dvipsnames]{color}
\usepackage{enumerate}
\usepackage{multirow}

\newcommand{\specialcell}[2][c]{%
  \begin{tabular}[#1]{@{}c@{}}#2\end{tabular}}

\usepackage{hyperref} %

\begin{document}
\title{Fisher Motion Descriptor\\ for Multiview Gait Recognition}

\cortext[cor]{Corresponding author \\%
Department of Computing and Numerical Analysis,\\
Escuela Polit\'ecnica Superior, Campus de Rabanales,\\
C\'ordoba, Spain, Tlf. 0034 957218980}

\address[uma]{Department of Computer Architecture, University of Malaga, Spain, 29071}
\address[uco]{Department of Computing and Numerical Analysis, University of Cordoba, Spain, 14071}

\author[uma]{F.M.~Castro}
\ead{fcastro@uma.es}

\author[uco]{M.J.~Mar\'in-Jim\'enez \corref{cor}}
\ead{mjmarin@uco.es}

\author[uma]{N.~Guil}
\ead{nguil@uma.es}

\author[uco]{R.~Mu\~noz-Salinas}
\ead{rmsalinas@uco.es}

\journal{Technical Report -- University of Cordoba}

\begin{abstract}
The goal of this paper is to identify individuals by analyzing their gait.
Instead of using binary silhouettes as input data (as done in many previous works) we 
propose and evaluate the use of motion descriptors based on densely sampled short-term trajectories.
We take advantage of state-of-the-art people detectors to define custom spatial configurations of the descriptors around the target person,  
obtaining a rich representation of the gait motion.
The local motion features (described by the Divergence-Curl-Shear descriptor~\cite{jain2013cvpr}) extracted 
on the different spatial areas of the person are combined into a single high-level gait descriptor 
by using the Fisher Vector encoding~\cite{perronnin2010eccv}. The proposed approach, 
coined \textit{Pyramidal Fisher Motion}, is experimentally validated 
on `CASIA' dataset~\cite{yu2006casia} (parts B and C), `TUM GAID' dataset~\cite{tumDB}, `CMU MoBo' dataset~\cite{Gross2001mobo} and the recent `AVA Multiview Gait' dataset~\cite{lopez2014ava}. 
The results show that this new approach achieves state-of-the-art results in the problem of gait recognition,
allowing to recognize walking people from diverse viewpoints on single and multiple camera setups, wearing different clothes, carrying bags, walking at diverse speeds and not limited to straight walking paths.
\end{abstract}
\maketitle

\section{Introduction}

The term \textit{gait} refers to the way each person walks. Actually, humans are good recognizing 
people at a distance thanks to their gait~\cite{cutting1977gait}, what provides a good (non invasive) 
way to identify people without requiring their cooperation, 
in contrast to other biometric approaches as iris or fingerprint analysis.
Some potential applications are access control in special areas (e.g. military bases or 
governmental facilities) or smart video surveillance (e.g. bank offices or underground stations),
where it is crucial to identify potentially dangerous people without their cooperation. 
Although great effort has been put into this problem in recent years~\cite{hu2004survey}, 
it is still far from solved.

\begin{figure}[th]
\centering
\includegraphics[width=0.48 \textwidth]{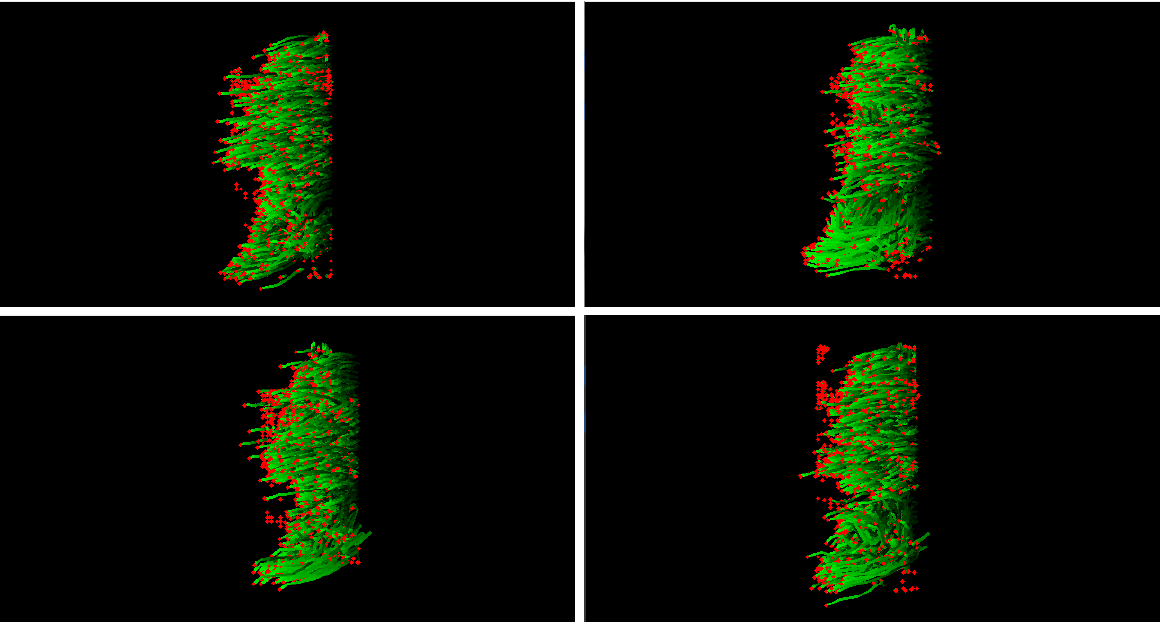}
\vspace{-0.01cm}
\caption{\textbf{Who are they?} The goal of this work is to identify people by using their gait. 
We build the Pyramidal Fisher Motion descriptor from trajectories of points. We represent here the gait motion of four different subjects.}
\label{fig:teaser}
\vspace{-0.05cm}
\end{figure}

Popular approaches for gait recognition require the computation of the binary silhouettes of 
people~\cite{han2006gei}, %
usually, by applying some background segmentation technique. 
However, this is a clear limitation in presence of dynamic backgrounds and/or non static cameras, 
where noisy segmentations are obtained.
To deal with these limitations, we propose the use of descriptors based on the local motion of points. 
These kind of descriptors have become recently popular in the field of human action recognition~\cite{wang2011cvpr}. 
The main idea is to build local motion descriptors from densely sampled points. Then, these local descriptors are aggregated into higher level descriptors by using histogram-based techniques (e.g. Bag of Words~\cite{Sivic03}).

Therefore, our research question is: \textit{could we identify people by using only local motion 
features as represented in Fig.~\ref{fig:teaser}?} We represent in Fig.~\ref{fig:teaser} the local 
trajectories (in green) of image points (in red) belonging to four different people. Our goal is to use each set of local 
trajectories (or tracklets) to build a high-level descriptor that allows to identify individuals. In this paper we 
introduce a new gait descriptor, coined \textit{Pyramidal Fisher Vector}, that combines the potential 
of recent human action recognition descriptors with the rich representation provided by 
Fisher Vectors encoding~\cite{perronnin2010eccv}. 
A thorough experimental evaluation is carried out on `CASIA Gait' dataset (sets B and C), `CMU MoBo' dataset, `TUM GAID' dataset and the recent `AVA Multiview Gait' dataset. The variety of the employed datasets will allow us to show the robustness of our gait recognition method in presence of challenging situations: poor silhouette segmentation, occlusion of body parts, strong changes in body scale, complex subject trajectories and changes in clothing. 
And the most important point, we will show that a discriminative classifier based on our features can handle multiple
viewpoints at test time, removing the limitation of using several camera viewpoints simultaneously,
even on curved walking paths.
Comparison with previous works has also been performed, showing that our approach outperforms previous techniques in most of scenarios, demonstrating that the new paradigm can compete successfully with current state-of-the-art gait recognition methods.

This paper is organized as follows. After presenting the related work, we describe our proposed
framework for gait recognition in Sec.~\ref{sec:methods}. Sections \ref{sec:expers}--\ref{sec:experesTUM} are devoted to the 
experimental results. 
An overall discussion of the results is expounded in Sec.~\ref{sec:discussion}
And, finally, the conclusions are presented in 
Sec.~\ref{sec:conclusions}.

\subsection{Related work}\label{subsec:relwork}
Many research papers have been published in recent years tackling the problem of human gait 
recognition using different sources of data like inertial sensors~\cite{Ngo20151289,Ngo2014228}, foot pressure~ \cite{Zheng20123603}, infrared images~\cite{Xue20102904} or the traditional images. For example, in \cite{hu2004survey} we can find a survey on this problem summarizing 
some of the most popular approaches. Some of them use explicit geometrical models of human bodies, 
whereas others use only image features. A sequence of binary silhouettes of the body is adopted in 
many works as input data.
In this sense, the most popular silhouette-based gait descriptor is the called Gait Enery Image 
(GEI)~\cite{han2006gei}. The key idea is to compute a temporal averaging of the binary silhouette 
of the target subject.
Liu et al.~\cite{liu2012icpr}, to improve the gait recognition performance, propose the computation 
of HOG descriptors from popular gait descriptors as the GEI and the Chrono-Gait Image (CGI).
In \cite{martin2012icpr}, the authors try to find the minimum number of gait cycles needed to 
carry out a successful recognition by using the GEI descriptor. 
Martin-Felez and Xiang~\cite{martin2012eccv}~\cite{MartinFelez20143793}, using GEI as the basic gait descriptor, propose a 
new ranking model for gait recognition. This new formulation of the problem allows to leverage training data from different datasets, thus, improving the recognition performance.
In \cite{akae2012cvpr}, Akae et al. propose a temporal super resolution approach to deal with 
low frame-rate videos for gait recognition. They achieve impressive results by using binary 
silhouettes of people at a rate of 1-fps.
Hu proposes in ~\cite{hu2013} the use of a regularized local tensor discriminant analysis method 
with the Enhanced Gabor representation of the GEI. In addition, the same author defines 
in~\cite{hu2014} a method to identify camera viewpoints at test time from patch distribution features.
Recently, Lai et al.~\cite{lai2014} proposed a novel discriminant subspace learning
method (Sparse Bilinear Discriminant Analysis) that extends methods based on matrix-representation
discriminant analysis to sparse cases, obtaining competitive results on gait recognition.
In many works it is assumed that the target person follows a straight path, however, 
Iwashita et al.~\cite{iwashita2014} explicitly focus on curved trajectories. Although, curved trajectories
is not an specific goal in our paper, we show results of our proposed method on unconstrained trajectory
paths, highlighting that the kind of trajectory is not a limitation for our proposal.

On the other hand, human action recognition (HAR) is related to gait recognition in the sense 
that the former also focuses on human motion, but tries to categorize such motion into categories of actions as \textit{walking, jumping, boxing}, etc. In HAR, the work of Wang et al.~\cite{wang2011cvpr} is a key reference. They introduce the use of short-term trajectories of densely sampled points for describing human actions, obtaining state-of-the-art results in the HAR problem. The dense trajectories are described with the Motion Boundary Histogram. Then, they describe the video sequence by using the Bag of Words (BOW) model~\cite{Sivic03}. Finally, they use a non-linear SVM with $\chi^2$-kernel for classification.
In parallel, Perronnin and Dance~\cite{perronnin2007cvpr} introduced a new way of histogram-based 
encoding for sets of local descriptors for image categorization: the Fisher Vector (FV) encoding. 
In FV, instead of just counting the number of occurrences of a visual word (i.e. quantized local descriptor) as in BOW, the concatenation of gradient vectors of a Gaussian Mixture is used. Thus, obtaining a larger but richer representation of the image.

Borrowing ideas from the HAR and the image categorization communities, we propose in this paper 
a new approach for gait recognition that combines low-level motion descriptors, extracted from 
short-term point trajectories, with a multi-level gait encoding based on Fisher Vectors: 
the \textit{Pyramidal Fisher Motion} (PFM) gait descriptor.
In this context, the work of Gong et al.~\cite{gong2013fisher}
is similar to ours in the sense that they propose a method that uses dense local spatio-temporal 
features and a Fisher-based representation rearranged as tensors. However, there are some 
significant differences: \textit{(i)} instead of dealing with a single camera viewpoint, 
we integrate in our system several camera viewpoints; \textit{(ii)} instead of using all the local features available in the 
sequence, we use a person detector to focus only on the ones related to the target subject; and, 
\textit{(iii)} the information provided by the person detector enables a richer representation 
by including coarse geometrical information through a spatial grid defined on the person 
bounding-box. 

A conference version of this paper was presented in~\cite{castro2014icpr}. 
In this current version, three important improvements have been introduced: (\textit{i}) we use a new person detector that employs a combination of full-body and upper-body detectors, resulting in a robust tool when occlusions happen in body parts; (\textit{ii}) to recover broken tracks of detections, we use a histogram-based linking process, obtaining longer tracks that allow a better representation of the gait; %
and, (\textit{iii}) three new databases have been included in the experimental results section: `CMU Motion of Body' dataset, the `CASIA Gait' dataset and the `TUM GAID' dataset.

On this set of databases we have conducted a variety of significant experiments to evaluate the robustness of our method: \textit{(i)} to evaluate the minimum number of frames necessary to identify a subject at test time ; \textit{(ii)} to take advantage of information of multiple cameras to build a classifier able to identify subjects under multiple views; and, \textit{(iii)} to identify subjects by employing their upper or lower body. Moreover, we have designed specific experiments to carry out a thorough comparison with previous methods and to conclude that our method outperforms state-of-the-art gait recognition approaches under most of scenarios.

\begin{figure*}[th]
\centering
 \includegraphics[width=0.98 \textwidth]{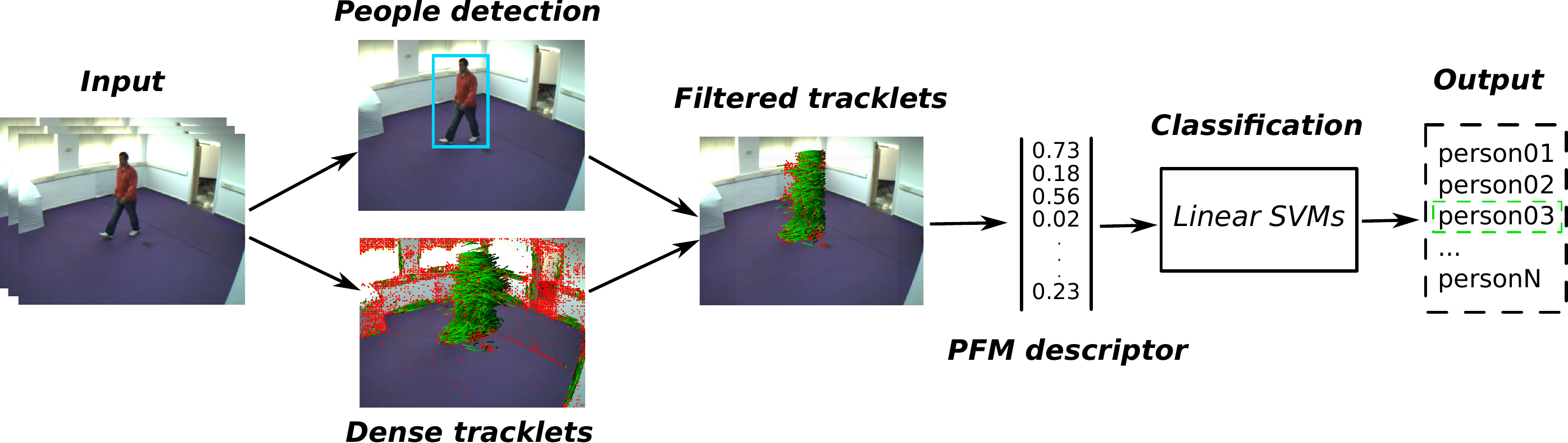}
\vspace{-0.01cm}
 \caption{\textbf{Pipeline for gait recognition.} Steps: a) The input is a sequence of video frames. b)
 Densely sampled points are tracked. People detection helps to remove trajectories not related to gait. 
 c) The PFM descriptor is computed on the filtered tracklets. d) The PFM descriptor is the input of a discriminative classifier to output a subject identity.
 }
\label{fig:pipeline}
\vspace{-0.05cm}
\end{figure*}
\section{Proposed framework}\label{sec:methods}

In this section we present our proposed framework to address the problem of gait recognition. 
Fig.~\ref{fig:pipeline} summarizes the pipeline of our approach.
We start by computing local motion descriptors from tracklets of densely sampled points on the whole 
scene (Sec.~\ref{subsec:features}). Since, we do not assume a static background, 
we run a person detector to remove the point trajectories that
are not related to people (Sec.~\ref{subsec:detection}). 
In addition, we spatially divide the person regions to aggregate the local motion descriptors into 
mid-level descriptors (Sec.~\ref{subsec:fv}). 
Finally, a discriminative classifier is used to identify the subjects (Sec.~\ref{subsec:svm}).

\subsection{Motion-based features}\label{subsec:features}
The first step of our pipeline is to compute densely sampled trajectories. Those trajectories are computed by following the approach of Wang et al.~\cite{wang2011cvpr}. 
Firstly, dense optical flow $F = (u_t,v_t)$ is computed~\cite{Farneback03} on a dense grid (i.e. step size of 5 pixels and over 8 scales). Then, each point $p_t = (x_t, y_t)$ at frame $t$ is tracked to the next frame by
median filtering as follows:
\begin{equation}
   p_{t+1} = (x_{t+1}, y_{t+1}) = (x_t, y_t) + (M * F)|_{(\bar{x_t},\bar{y_t})} 
\end{equation}
where $M$ is the kernel of median filtering and $(\bar{x_t},\bar{y_t})$ is
the rounded position of $p_t$. To minimize drifting effect, the tracking is limited to $L$ frames. We use $L=15$ as in~\cite{jain2013cvpr}. As a postprocessing step, noisy and uninformative trajectories (e.g. excessively short or showing sudden large displacements) are removed. These short-term trajectories (or tracklets) are represented in Fig.~\ref{fig:pipeline} by green lines for each considered point (in red).

Once the local trajectories are computed, they are described with the Divergence-Curl-Shear (DCS) descriptor proposed by Jain et al.~\cite{jain2013cvpr}, which is computed as follows:

\begin{equation}\label{eq:DCS}
\left\{
\begin{aligned}
	\text{div}(p_t)  &=& \frac{\partial u(p_t)}{\partial x} + \frac{\partial v(p_t)}{\partial y} \quad \\
	\text{curl}(p_t) &=& \frac{-\partial u(p_t)}{\partial y} + \frac{\partial v(p_t)}{\partial x} \quad \\
	\text{hyp}_1(p_t) &=& \frac{\partial u(p_t)}{\partial x} - \frac{\partial v(p_t)}{\partial y} \quad \\
	\text{hyp}_2(p_t) &=& \frac{\partial u(p_t)}{\partial y} + \frac{\partial v(p_t)}{\partial x} \quad \\	
\end{aligned}
\right.
\end{equation}

As described in \cite{jain2013cvpr}, the divergence is related to axial motion, expansion and scaling effects, whereas the curl is related to rotation in the image plane. From the hyperbolic terms ($\text{hyp}_1,\text{hyp}_2$), we can compute the magnitude of the shear as:

\begin{equation}\label{eq:shear}
  \text{shear}(p_t) = \sqrt{\text{hyp}_1^2(p_t) + \text{hyp}_2^2(p_t)}
\end{equation}

Then, those kinematic features are combined in pairs as in \cite{jain2013cvpr} to get the final motion descriptors. 

\subsection{People detection and tracking}\label{subsec:detection}
We follow a tracking-by-detection strategy as in~\cite{Eichner2012ijcv}: we detect people with the detection framework of Felzenszwalb et al.~\cite{felzenszwalb10pami}; and, then, we apply the clique partitioning algorithm of Ferrari et al.~\cite{Ferrari01} to group detections into tracks. 
Short tracks with low-scored detections are considered as false positives and are discarded for further processing. 
We are able to recover broken tracks by linking non-overlapping (in time) tracks that are similar enough 
based on histograms of color (i.e. $3\times16$-bins histograms on RGB space and $\chi^2$ distance). To obtain the color histogram of a track, we compute the average of all histograms of each bounding-box that compose a track. Finally, we compare the similarity of tracks through $\chi^2$ distance. This is especially useful when the detector misses the person for a period of time.
In addition, to remove false positives generated by static objects, we measure the displacement of the detection along the sequence. Thus, discarding those tracks showing a static behaviour.

The person tracks finally kept are used to filter out the trajectories that are not related to people: 
we only keep the tracklets that pass through, at least, one bounding-box of any person track. 
In this way, we can focus on the trajectories that should contain information about the gait.

\begin{figure}[t]
\centering
\includegraphics[width=0.48 \textwidth]{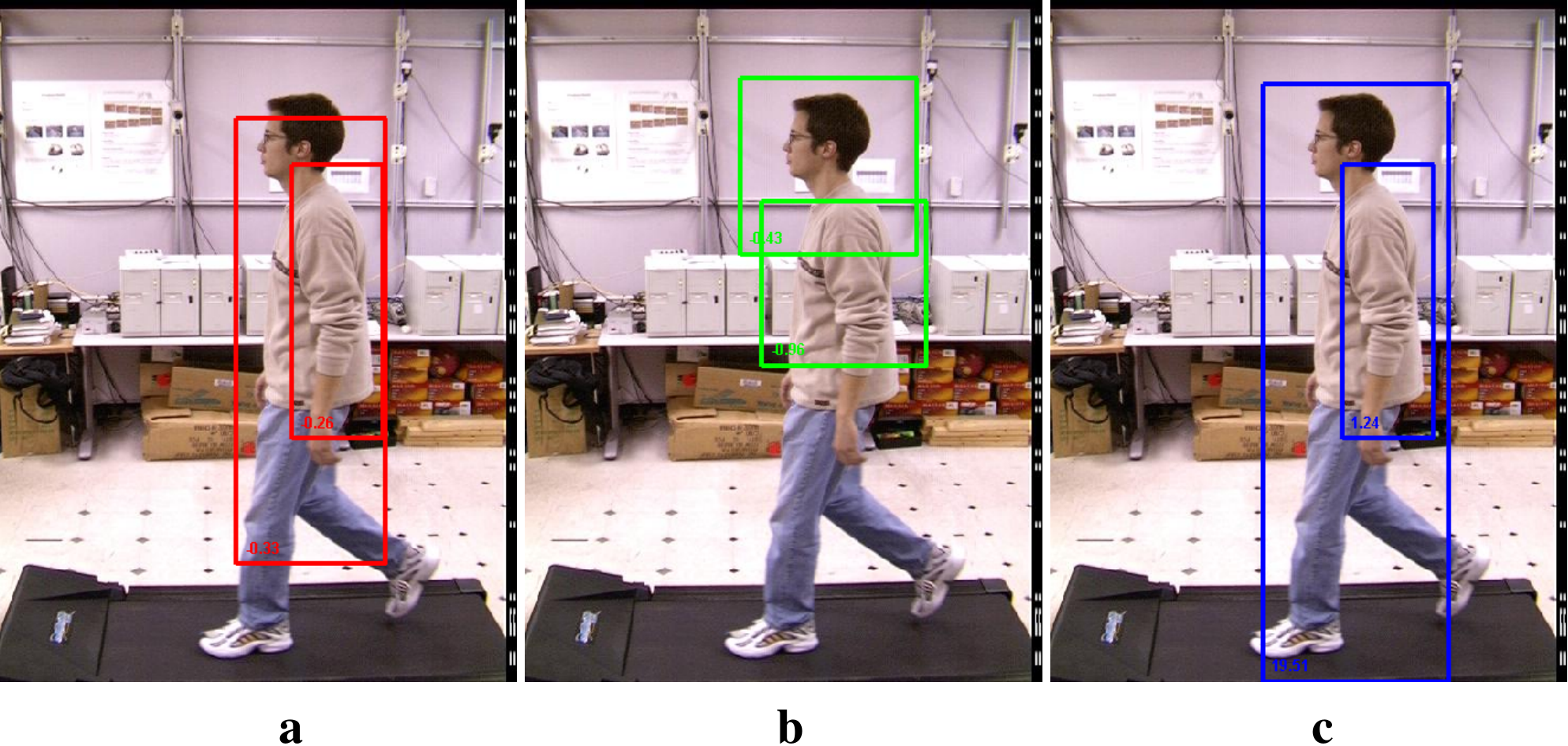}
\vspace{-0.01cm}
\caption{\textbf{Combining full-body and upper-body detections.} \textbf{(a)} Top 2 detections returned by the full-body detector. Note how none of them covers the full height of the person.  \textbf{(b)} Top 2 detections returned by the upper-body detector. The highest scored BB will lead to the expected full-body BB after the geometrical transformation into full-body BB. 
\textbf{(c)} Final BBs obtained after the combination of the detections in (a) and (b) and non-maxima suppression. 
Note how the BBs are re-scored during the combination procedure. (Best viewed on pdf)}
\label{fig:combFBUB}
\vspace{-0.05cm}
\end{figure}

\noindent \textbf{People detection.} %
On each video frame, we run both a pedestrian detector~\cite{felzenszwalb10pami} (i.e. full body) 
and an 
upper-body detector~\cite{marin2013ijcv}. Since the pedestrian detector favors the detection of 
people standing with the legs at rest, the idea of using the upper-body detector is to be able to 
detect people holding poses not covered by the pedestrian detector (see Fig.~\ref{fig:combFBUB}.a), 
or people with the legs partially occluded.

Inspired by the work of Kl\"aser~\cite{klaser06master}, after running both 
detectors (Fig.~\ref{fig:combFBUB}.a--b), we transform the upper-body bounding-boxes (BB) into 
pedestrian-like BB.

Given an upper-body BB $(x,y,w,h)$ with center $(x,y)$, width $w$ and height $h$, its transformed 
BB $(x',y',w',h')$ is given by the following equations:
$$
  x' = x + \mu_x \cdot h ;
  y' = y + \mu_y \cdot h ;
  w' = \mu_w  \cdot w  ;
  h' = \mu_h \cdot h  
$$

Where $\mu_x$, $\mu_y$, $\mu_w$ and $\mu_h$ are parameters learnt from training samples as the mean 
value of relative locations ($x_r$, $y_r$) and scales ($w_r$, $h_r$).
In particular, we use the following relations:
$$
x_r = \frac{x_f - x_u}{h_u} ;
y_r = \frac{y_f - y_u}{h_u} ;
w_r = w_f / w_u;
h_r = h_f /h_u
$$
Where $x_r$ and $y_r$ are the scale-invariant relative coordinates between the center of a 
full-body BB  $(x_f,y_f,w_f,h_f)$ and the center of an upper-body BB $(x_u,y_u,w_u,h_u)$; and $w_r$  
and $h_r$ are the relative widths and heights of the BBs. 
Since we have now two classes of detections, upper-bodies (UB) and full-bodies (FB), it is more than likely that an UB is found inside a FB. Therefore, we define a procedure similar to the Kl\"aser's one~\cite{klaser06master} to combine UB and FB detections. Firstly, all the UB detections are geometrically transformed into FBs, and the detection scores of both classes are scaled to the same range to make them comparable. 
Then, for each FB detection, (\textit{i}) we select from all the unused UBs the one that overlaps more (in terms of intersection-over-union, IoU) with the current FB detection (recall that the geometrical transformation has been previously applied to the UB); (\textit{ii}) if the IoU is greater than a combining threshold $\tau_C$, then, the selected UB is marked as `used' and a new BB is added to the new detection set $\mathcal{S}$, where such BB is defined as the one with the largest area from the FB and the transformed UB; (\textit{iii}) if a combination has been done, we assign a new score $S_c$ to the resulting BB as $S_c = S_f \cdot S_u \cdot IoU$, where $S_f$ and $S_u$ are the original detection scores of the FB and the UB, respectively.
Finally, the unused UBs (transformed) and the non combined FBs are included into set $\mathcal{S}$.

Note that more than one BB in $\mathcal{S}$ could now cover a significant part of the same image region, therefore, a non-maxima-suppression (NMS) procedure is applied to the new set of BBs to obtain the subset of BBs that will be used for further processing. An example of the resulting BBs is shown in Fig.~\ref{fig:combFBUB}.c, where from four window detections we end up with only two final BBs with FB-like aspect-ratio and new scores.

\subsection{Pyramidal Fisher Motion Descriptor}\label{subsec:fv}
\noindent \textbf{Fisher Motion.} %
As described above, our low-level features are based on motion properties extracted from person-related local trajectories. In order to build a person-level gait descriptor, we need to summarize the local features. We propose here the use of Fisher Vectors (FV) encoding~\cite{perronnin2010eccv}.

The FV, that can be seen as an extension of the Bag of Words (BOW) representation~\cite{Sivic03}, builds on top of a Gaussian Mixture Model (GMM), where each Gaussian corresponds to a visual word. Whereas in BOW, an image is represented by the number of occurrences of each visual word, in FV an image is described by a gradient vector computed from a generative probabilistic model.
The dimensionality of FV is $2ND$, where $N$ is the number of Gaussians in the GMM, and $D$ is the dimensionality of the local motion descriptors $x_t$. For example, in our case, the dimensionality of the local motion descriptors is $D=318$\footnote{%
2D coordinates: 30; Div+Curl: 96; Curl+Shear: 96; Div+Shear: 96 
}%
, if we use $N=100$ Gaussians, then, the FV would have 63600 dimensions.
In this paper, we will use the term \textit{Fisher Motion} (FM) to refer to the FV computed on a video from low-level motion features.

Assuming that our local motion descriptors $\{x_t \in R^D,t=1 \ldots T\}$ of a video $V$ are generated independently by a GMM $p(x|\lambda)$ with parameters $\lambda=\{w_i, \mu_i, \Sigma_i , i=1 \ldots N\}$, we can represent $V$ by the following gradient vector~\cite{perronnin2007cvpr}:
\begin{equation}\label{eq:fv}
G_\lambda (V) = \frac{1}{T} \sum_{t=1}^{T}{\nabla_\lambda \log p(x_t|\lambda)}
\end{equation}
\noindent where $T$ is the total number of local descriptors %
and $\nabla_\lambda$ denotes the gradient operator with respect to $\lambda$.

Following the proposal of~\cite{perronnin2010eccv}, to compare two videos $V$ and $W$, a natural kernel on these gradients is the Fisher Kernel: $K(V,W) = G_\lambda(V)^T F_\lambda^{-1} G_\lambda(W)$, where $F_\lambda$ is the Fisher Information Matrix. %
As $F_\lambda$ is symmetric and positive definite, it has a Cholesky decomposition $F_\lambda^{-1}=L_\lambda^T L_\lambda$, and 
$K(V,W)$ can be rewritten as a dot-product between normalized vectors $\Gamma_\lambda$ with: $\Gamma_\lambda(V)=L_\lambda G_\lambda(V)$. Then, $\Gamma_\lambda(V)$ is known as the Fisher Vector of video V.
As stated in \cite{perronnin2010eccv}, the capability of description of the FV can be improved by applying it a signed square-root followed by L2 normalization. So, we adopt this finding for our descriptor.
\noindent \textbf{Pyramidal representation.} %
We borrow from \cite{Marin2012paaa} the idea of building a pyramidal representation of the gait motion.
Since each bounding-box covers the whole body of a single person, we propose to spatially divide the BB into cells. 
Then, a Fisher vector is computed inside each cell of the spatio-temporal grid. 
We can build a pyramidal representation by combining different grid configurations. Then, the final feature vector, %
used to represent a time interval, is computed as the concatenation of the cell-level Fisher vectors from all the levels of the pyramid. 
\begin{figure*}[th]
\begin{center}
  \includegraphics[width=0.96 \textwidth]{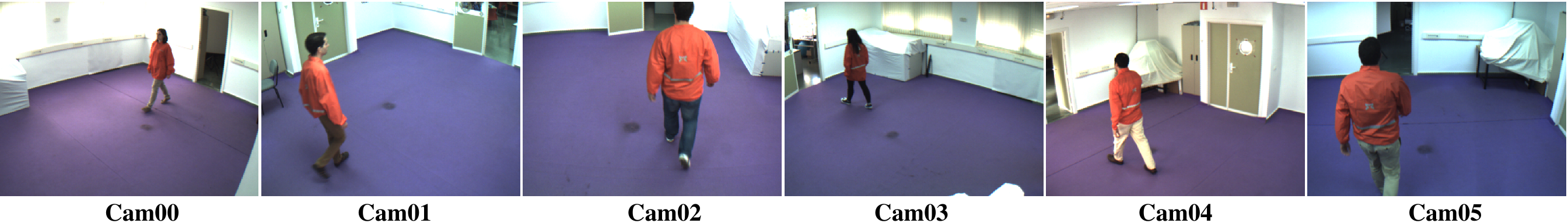}%
\vspace{-0.01cm}	
 \caption{\textbf{AVAMVG dataset.} Different people recorded from six camera viewpoints. The dataset 
contains both female and male subjects performing different trajectories through the indoor 
scenario. Note that cameras \textit{Cam02} and \textit{Cam05} are prone to show people 
partially occluded.}
 \label{fig:dataset}
\vspace{-0.05cm}
\end{center}
\end{figure*}
\subsection{Classification}\label{subsec:svm}
The last stage of our pipeline is to train a discriminative classifier to distinguish between the different human gaits. 
Since, this is a multiclass problem, we train $P$ binary Support Vector Machines 
(SVM)~\cite{osuna1997svm} (as many as different people) in a \textit{one-vs-all} strategy. 
Although the $\chi^2$ kernel %
is a popular choice for BOW-based descriptors, a linear kernel is typically enough for FV, due to the rich feature representation that it provides. 
\subsection{Implementation details}\label{subsec:code}
For people detection, we use the code published by the authors of \cite{felzenszwalb10pami}.
For computing the local motion features, we use the code published by 
the authors of \cite{jain2013cvpr}. The Fisher Vector encoding and the classification is carried out 
by using the code included in the library VLFeat\footnote{%
VLFeat library is available at \url{http://www.vlfeat.org/}
}.%

\section{Overview of the experiments}\label{sec:expers}
In order to validate our approach, we carry out diverse experiments on four datasets: 
``AVA Multi-View Dataset'', ``CMU MoBo Dataset'', ``CASIA B and C Datasets'' and ``TUM GAID Dataset''.
With these experiments we try to answer, among others, the following questions: 
 a) \textit{is the combination of trajectory-based features with FV a valid approach for gait recognition?}; 
 b) \textit{can we learn different camera viewpoints in a single classifier?}; 
 c) \textit{can we improve the recognition rate by spatially dividing the human body region?}; 
 d) \textit{what is the effect of using PCA-based dimensionality reduction on the recognition performance?}; 
 e) \textit{what is the influence of the sequence length in the recognition performance?}; 
 f) \textit{is it necessary to use the DCS descriptor as a whole or can we use just some of its components?}; and, 
 g) \textit{can the proposed model generalize well on unrestricted walk trajectories?} 

The subsequent sections present the experiments and results obtained on the different datasets.

\section{Experimental results on AVAMVG}\label{sec:experesAVA}
The first dataset where we perform our experiments is the ``AVA Multi-View Dataset for Gait Recognition'' (AVAMVG)~\cite{lopez2014ava}. %
In AVAMVG 20 subjects perform 10 walking trajectories in an indoor environment. 
Each trajectory is recorded by 6 color cameras placed around a room that is crossed by the subjects during the performance.
Fig.~\ref{fig:dataset} shows the scenario from the six available camera viewpoints. 
Note that depending on the viewpoint and performed trajectory, people appear at diverse scales, 
even showing partially occluded body parts. In particular, the 3rd and 6th camera viewpoints 
represented in Fig.~\ref{fig:dataset} are more likely to show partially visible bodies most of the 
time than the other four cameras. Therefore, in our experiments, and without loss of generality, 
we will use only four cameras (i.e. \textit{Cam00,Cam01,Cam03,Cam04}).
Trajectories 1 to 3 follow a linear path, whereas the remaining seven trajectories are curved. 
The released videos have a resolution of $640 \times 480$ pixels. Each video has around 375 frames, 
where only approximately one third %
of the frames contains visible people.
\subsection{Experimental setup}\label{subsec:expersetup}
\begin{table*}[t] %
\renewcommand{\arraystretch}{1.0}
\caption{\textbf{Recognition results on AVAMVG: experiments A, B and C.} Each entry contains the percentage of 
correct recognition in the multiview setup and, in parenthesis, the recognition per single view. 
Each row corresponds to a different configuration of the gait descriptor. 
$K$ is the dictionary size. Best results are marked in bold.
(See main text for details)}
\label{tab:resultsA}
\centering
\footnotesize
\begin{tabular}{l|c| c c c c}
 \hline
 \textit{Experiment} & $K$ & \textit{Trj=1+2} & \textit{Trj=1+3} & \textit{Trj=2+3} & \textit{Avg}\\
 \hline
  BOW       & 4000 & 95 (78.8) & 85 (62.5) & 100 (84.4) & 93.3 (75.2) \\ %
  PFM        & 150 & 100 (98.8) & 100 (96.2)  & 100 (97.5) & 100 (97.5) \\
  PFM-FB    & 150  & 100 (98.8)  &  100 (95)  & 100 (100) & 100 (97.9) \\ %
  PFM-H1     & 150 & 100 (95)  & 100 (87.5)   & 100 (97.5) & 100 (93.3) \\
  PFM-H2     & 150 & 100 (97.5) & 95 (93.8)   & 100 (97.5) & 98.3 (96.3) \\
  PFM+PCAL50 & 150 & 100 (100) & 100 (97.5)  & 100 (98.8)  & 100 (98.8) \\
  PFM+PCAH256& 100 & 100 (100)  & 100 (97.5)   & 100 (98.8) & 100 (98.8)\\
  PFM+PCAL100+PCAH256& 150 & 100 (100) & 100 (97.5) & 100 (98.8) & 100 (98.8)\\ %
  \hline
  PFM+PCAL50+PCAH256+pyr& 100 & 100 (100) & 100 (97.5) & 100 (98.8) & \textbf{100 (98.8)}\\ %
 \hline
\end{tabular}
\vspace{-0.02cm}
\end{table*}
Since we have multiple viewpoints of each \textit{instance} (i.e. pair subject--trajectory), we can assign a single label to it by \textit{majority voting} on the viewpoints. This approach helps to deal with labels wrongly assigned to individual viewpoints. Note that instead of training an independent classifier (see Sec.~\ref{subsec:svm}) per camera viewpoint, we train a single classifier with samples obtained from different camera viewpoints, allowing the classifier to learn the relevant gait features of each subject from multiple viewpoints.
In order to increase the amount of training samples, we generate their \textit{mirror} sequences, thus, 
doubling the amount of samples available during learning. 

We describe below the different experiments performed to give
answer to the questions stated at Sec.~\ref{sec:expers}. 
After the description of the experiments, the achieved results are discussed in Sec.~\ref{subsec:resultsAVA}.

\noindent \textbf{Experiment A: baseline.} %
We use the popular Bag of Words approach (BOW)~\cite{Sivic03} as baseline, which is compared to our approach. For this experiment, we use trajectories 1, 2 and 3 (i.e. straight path). We use a leave-one-out strategy on the trajectories (i.e. two for training and one for test). We sample dictionary sizes in the interval $[500,4000]$ for BOW\footnote{%
Larger dictionary sizes for BOW did not show any significative improvement. In contrast, the computational time increased enormously. 
}%
, and in the interval $[50,200]$ for PFM. Both BOW and PFMs have a single level with two rows and one column (i.e. concatenation of two descriptors: half upper-body and half lower-body). 
The results of this experiment are included in Tab.~\ref{tab:resultsA}: rows `BOW' and `PFM'.

\noindent \textbf{Experiment B: half body features.} %
Focusing on PFM, we compare in Tab.~\ref{tab:resultsA} four configurations of the PFM 
on trajectories 1, 2 and 3: a) no spatial partition of the body (row `PFM-FB'); b) using only the 
top half of the body (row `PFM-H1'); c) using only the bottom half of the body (row `PFM-H2'); and, 
d) using the concatenation of the top and bottom half of the body (row `PFM'). 

\noindent \textbf{Experiment C: dimensionality reduction.} %
Since the dimensionality of PFM is typically large, we evaluate in this experiment the impact of 
dimensionality reduction on the final recognition performance. We run Principal Component Analysis 
(PCA) both on the original low-level features (318 dimensions), and on the PFM vectors. We use the 
PFM descriptor, as in experiment A, on trajectories 1, 2 and 3.
The results of this experiment are included in Tab.~\ref{tab:resultsA}: row `PFM+PCAL50' indicates 
that the low-level feature vectors have been compressed to 50 dimensions; row `PFM+PCAH256' 
indicates that the final PFM feature vectors have been compressed to 256 dimensions; row 
`PFM+PCAL100+PCAH256' means that the low-level features have been intially compressed to 100 
dimensions and the final PFM vectors to 256 dimensions; and, finally, row `PFM+PCAL50+PCAH256+pyr' 
corresponds to a two level pyramid with the compression indicated in the name.

\begin{table}[t] %
\renewcommand{\arraystretch}{1.2}
\setlength{\tabcolsep}{5pt}
\caption{\textbf{Recognition results on AVAMVG: experiment D.} Each 
entry contains the percentage of correct recognition in the multiview setup and, in parenthesis, 
the recognition per single view. Each row corresponds to a different length of the test sequence. 
$K$ is the GMM size used for FM. Best results are marked in bold.
(See main text for further details.)}
\label{tab:resultsMinLen}
\centering
\scriptsize
\begin{tabular}{l|c| c c c c}
 \hline
 \textit{Experiment} & $K$ & \textit{Trj=1+2} & \textit{Trj=1+3} & \textit{Trj=2+3} & \textit{Avg}\\
 \hline
  PFM-len15        & 150 & 100 (87.5) & 95.0 (83.8)  & 100 (87.5) & 98.3 (86.3) \\
  PFM-len20        & 150 & 100 (88.8) & 95.0 (90.0)  & 100 (93.8) & 98.3 (90.9) \\
  PFM-len25        & 150 & 100 (90.0) & 90.0 (91.3)  & 100 (95.0) & 96.7 (92.1) \\
  PFM-len30        & 150 & 100 (91.3) & 100 (86.3)  & 100 (92.5) & 100 (90.0)\\
  PFM-len40        & 150 & 100 (96.3) & 100 (91.3)  & 100 (96.3) & 100 (94.6) \\
  PFM-len50        & 150 & 100 (93.8) & 100 (96.3)  & 100 (98.8) & \textbf{100 (96.3)} \\
 \hline
\end{tabular}
\vspace{-0.06cm}
\end{table}

\noindent \textbf{Experiment D: influence of sequence length.} %
The goal of this experiment is to evaluate the influence of the sequence 
length in the recognition process. For this purpose, at test time, we use only a single subsequence 
of $T$ frames\footnote{%
Note that a buffer of $L=15$ video frames is needed to compute the set of 
dense trajectories. Therefore, a value of $+14$ actual video frames has to be added to $T$. For example,
$T'=15+14=29$, around one second.  
}%
extracted around the middle of the sequence. 
In Tab.~\ref{tab:resultsMinLen}, each row corresponds to a different number 
of frames in the range $[15,50]$. The dictionary size has been fixed to $150$ components.

\noindent \textbf{Experiment E: testing on a single camera.} %
The previous results assume a multicamera enviroment at test time 
to combine information from diverse viewpoints. However, in many real situations, only a viewpoint 
is available at test time. In this experiment we present a breakdown of the recognition rates per 
camera.
The results of this experiment are summarized in Tab.\ref{tab:resultsCam}. 
We use here the configuration `PFM-len50'  learnt during the previous experiment (see 
Tab.~\ref{tab:resultsMinLen}). Each row corresponds to a different camera viewpoint (see 
Fig.~\ref{fig:dataset}).

\begin{table}[t] %
\renewcommand{\arraystretch}{1.1} %
\caption{\textbf{Recognition results on AVAMVG: experiment E.} 
Each entry contains the percentage of correct recognition for the given camera and training--test split. 
Each row corresponds to a different camera viewpoint. 
$K$ is the GMM size used for FM. Best results are marked in bold.
(See main text for details)}
\label{tab:resultsCam}
\centering
\scriptsize
\begin{tabular}{l|c| c c c c}
 \hline
 \textit{Experiment} & $K$ & \textit{Trj=1+2} & \textit{Trj=1+3} & \textit{Trj=2+3} & \textit{Avg}\\
 \hline
  PFM-len50-\textit{Cam00}       & 150 & 90  &  100 & 100 & 96.7 \\
  PFM-len50-\textit{Cam01}       & 150 & 90  &  90  & 95  & 91.7 \\
  PFM-len50-\textit{Cam03}       & 150 & 100 &  100 & 100 & \textbf{100} \\
  PFM-len50-\textit{Cam04}       & 150 & 95  &  95  & 100 & 96.7 \\
  \hline
\end{tabular}
\vspace{-0.02cm}
\end{table}

\noindent \textbf{Experiment F: feature selection.} %
In the previous experiments, our local motion vectors had 318 dimensions, 
obtained as the concatenation of four kind of features (see Sec.~\ref{subsec:features}): normalized 
coordinates, Div+Curl (DC), Curl+Shear (CS) and Div+Shear (DS). The goal of this experiment is to evaluate the contribution of each type of local feature in the gait recognition process. In this case, we learn independent 
dictionaries per feature type, instead of a single dictionary of concatenated local features as done 
before. Then, we concatenate the resulting feature-specific FVs. 
Tab.~\ref{tab:resultsFeatDics} summarizes the results of this experiment. 
Each row corresponds to a different configuration of the PFM descriptor, where `ft\textit{xxxx}' 
indicates the set of features used, encoded in the \textit{xxxx} string as follows (from left to 
right): normalized coordinates, DC, CS and DS. Value $1$ means \textit{used} whereas value 
$0$ means \textit{not used}. For example, string `ft0110' means that only DC and CS descriptors 
have been used.

\noindent \textbf{Experiment G: training on straight paths and testing on curved paths.} %
In this experiment, we use the PFM descriptor as in experiment F. 
We use trajectories 1 to 3 for training, and trajectories 4 to 10 for testing. Note that in the 
latter sequences, the subjects perform curved trajectories, thus, changing their viewpoint (with 
regard to a given camera). 
The results of this experiment are summarized in Tab.~\ref{tab:resultsTrX}, where each column 
correspond to a different test trajectory. For each trajectory, 20 multiview sequences are 
evaluated (one per subject), corresponding to 80 individual camera viewpoints\footnote{%
Trajectory \#08 is not available for subject \textit{rafa}; and, trajectory \#10 is not available for 
subject \textit{angel}.
}.%

\begin{table*}[th] %
\renewcommand{\arraystretch}{0.8}
\caption{\textbf{Recognition results  on AVAMVG: experiment F.} 
Each entry contains the percentage of correct recognition in the multiview setup and, in 
parenthesis, the recognition per single view. Each row corresponds to a different configuration of 
the gait descriptor. `ft\textit{xxxx}' indicates the selected subfeatures used to describe the dense trajectories.
$K$ is the GMM size used for FM. Best results are marked in bold.
(See main text for further details.)}
\label{tab:resultsFeatDics}
\centering
\scriptsize
\begin{tabular}{l|c| c c c c}
 \hline
 \textit{Experiment} & $K$ & \textit{Trj=1+2} & \textit{Trj=1+3} & \textit{Trj=2+3} & \textit{Avg}\\
 \hline
  PFM-ft1000       & 100 & 95 (86.3) &  90 (71.3)  & 95 (87.5) & 93.3 (81.7) \\
  PFM-ft0100       & 100 & 100 (97.5) & 100 (96.2) & 100 (97.5) & 100 (97.1) \\
  PFM-ft0010       & 100 & 95 (96.3) & 100 (92.5)  & 100 (97.5) & 98.3 (95.4) \\
  PFM-ft0001       & 100 & 100 (97.5) & 100 (92.6) & 100 (97.5) & 100 (95.9) \\
  
  PFM-ft0011       & 100 & 100 (96.2) & 100 (91.2) & 100 (97.5) & 100 (95) \\
  
  PFM-ft0101       & 100 & 100 (100) & 100 (95.0)  & 100 (98.8) & \textbf{100 (97.9)} \\
  PFM-ft0110       & 100 & 100 (100) & 100 (95.0)  & 100 (98.8) & \textbf{100 (97.9)} \\
  PFM-ft0111       & 100 & 100 (100) & 100 (95.0)  & 100 (98.8) & 100 (97.9) \\
  PFM-ft1111       & 100 & 100 (100) & 100 (95.0)  & 100 (98.8) & 100 (97.9) \\
  \hline
  PFM+PCALx10+PCAH256-ft0110 & 100 & 95 (95) & 100 (95)  & 100 (97.5) & 98.3 (95.8)  \\
  
  PFM+PCALx20+PCAH256-ft0110 & 100 & 100 (96.3) & 100 (93.8)  & 100 (97.5) & 100 (95.9) \\
  
  PFM+PCALx40+PCAH128-ft0110 & 50 & 100 (95.0) & 100 (91.3)  & 100 (96.3) & 100 (94.2) \\

  PFM+PCALx40+PCAH256-ft0110 & 100 & 100 (97.5) & 100 (96.3)  & 100 (98.8) & 100 (97.5) \\
  
  PFM+PCALx40+PCAH256-ft0111 & 100 & 100 (96.2) & 100 (97.5) & 100 (97.5) & 100 (97.1) \\
 \hline
\end{tabular}
\vspace{-0.06cm}
\end{table*}
\begin{table*}[th] %
\renewcommand{\arraystretch}{1.0}
\caption{\textbf{Recognition results on curved trajectories of AVAMVG.} Training 
on trajectories $1+2+3$. Each column indicates the tested trajectory and each row corresponds to a 
different configuration of the gait descriptor. $K$ is the GMM size used for FM. Best results are 
marked in bold. (See main text for further details.)}
\label{tab:resultsTrX}
\centering
\scriptsize
\setlength{\tabcolsep}{0.25em} %
\begin{tabular}{l|c| c c c c c c c}
 \hline
 \textit{Experiment} & $K$ & \textit{Test=04} & \textit{Test=05} & \textit{Test=06} & 
\textit{Test=07} & \textit{Test=08} & \textit{Test=09} & \textit{Test=10}\\
 \hline
  PFM-ft0110                 & 150 & 90 (91.2) & \textbf{95 (85.9)} & 95 (84.6) & 95 (93.8) & 94.7 (89.3) & 95 
(91.1) & 95 (89.9) \\ 
  PFM+PCALx80+PCAH256-ft0111 & 150 & 90 (93.8) & 95 (83.1) & \textbf{100 (87.3)} & \textbf{95 (96.2)} & 94.7 
(88) & 95 (92.4) & 90 (86.1) \\
  PFM+PCALx50+PCAH256-ft0111 & 150 & 90 (92.5) & 90 (84.5) & 95 (87.3) & 95 (93.8) & 94.7 (88) & 
95 (92.4) & 90 (86.1) \\
  PFM+PCALx40+PCAH256-ft0111 & 150 & \textbf{95 (93.8)} & 90 (83.1) & 95 (88.7) & 95 (95)
& \textbf{94.7 (90.7)} & 95 (93.7) & \textbf{95 (89.9)}  \\
  PFM+PCALx40+PCAH256-ft0110 & 150 & 95 (91.2) & 95 (84.5) & 95 (84.5) & 95 (93.8) & 94.7 (89.3) & 95 (92.4) & 90 (87.3)  \\
  PFM+PCALx10+PCAH256-ft0110 & 150 & 90 (83.8) & 85 (80.3) & 85 (76.9) & 90 (83.8) & 84.2 (76.3) & 
95 (78.5) & 90 (84.8)  \\
  \hline
  PFM+PCALx40+PCAH256+pyr-ft0111 & 150 & 95 (83.5) & 90 (83.1) & 95 (87.3) & 95 (93.8) & 94.7 (88) & \textbf{95 (94.9)} & 95 (84.8) \\
  PFM+PCALx40+PCAH256+pyr-ft0110 & 100 & 95 (91.2) & 90 (79.2) & 95 (78.9) & 95 (91.2) & 94.7 (86.7) 
& 95 (91.1) & 85 (84.8) \\ 
  PFM+PCALx30+PCAH256+pyr-ft0111 & 150 & 95 (91.2) & 95 (83.1) & 95 (88.7) & 95 (93.8) & 
94.7 (88) & 95 (93.7) & 90 (86.1) \\

 \hline
\end{tabular}
\end{table*}
\subsection{Results}\label{subsec:resultsAVA}

The results shown in Tab.~\ref{tab:resultsA} correspond to \textit{experiments A, B and C} (see Sec.~\ref{subsec:expersetup}) and have been obtained by training on two of the three  straight trajectories ($\{1,2,3\}$) and testing on the remaining one (e.g. `\textit{Trj=1+2}' indicates training on trajectories \#1 and \#2, then, testing on trajectory \#3).
Therefore, each model is trained with 160 samples (i.e. 20 subjects $\times$ 4 cameras  $\times$ 2 trajectories) 
and tested on 80 samples.
Each column `\textit{Trj=X+Y}' contains the percentage of correct recognition per partition at 
instance level (i.e. combining the four viewpoints by majority voting) and, in parenthesis, at video level
(i.e. supposing that each camera viewpoint is an independent sample); 
column `\textit{Avg}' contains the average on the three partitions. Column $K$ refers to the number of centroids used for quantizing the low-level features in each FM descriptor.
Row `BOW' corresponds to the baseline approach (see Sec.~\ref{subsec:expersetup}). 
Row `PFM-FB' corresponds to the PFM on the full body (no spatial partitions). 
Rows `PFM-H1'and `PFM-H2' correspond to PFM on the top half and on the bottom half of the body, respectively.
Row `PFM' corresponds to a single-level PFM obtained by the concatenation of the descriptors extracted from both the top and bottom half of the body.
Row `PFM+PCAL50' corresponds to our proposal but reducing with PCA the dimensionality of the low-level motion descriptors to 50 before building our PFM (i.e. final gait descriptor with $K=150$ is $15000$-dimensional).
Row `PFM+PCAH256' corresponds to our proposal but reducing with PCA the dimensionality of our final PFM descriptor to 256 dimensions before learning the classifiers (i.e. final gait descriptor is $256$-dimensional).
Row `PFM+PCAL100+PCAH256' corresponds to our proposal but reducing both the dimensionality of the low-level descriptors and the final PFM descriptor (i.e. final gait descriptor is $256$-dimensional). %
Row `PFM+PCAL50+PCAH256+pyr' corresponds to a two-level pyramidal configuration where the first level has no spatial partitions and the second level is obtained by dividing the bounding box in two parts along the vertical axis, as done previously. In addition, PCA is applied to both the low-level descriptors and the final PFM vector.

The results shown in Tab.~\ref{tab:resultsTrX} correspond 
to \textit{experiment D} (see Sec.~\ref{subsec:expersetup}) and have been obtained by training on 
trajectories $\{1, 2, 3\}$ (all in the same set), and testing on trajectories $\{4,5,6,7,8,9,10\}$ 
(see corresponding columns). As done in the previous experiments, different configurations of PFM 
have been evaluated. Each entry of the table contains the percentage of correct recognition in the 
multiview setup and, in parenthesis, the recognition per video. 
From the seven tested trajectories, only on trajectory number \#06 a perfect recognition rate was 
achieved on the multiview setup. 
In contrast, the trajectory number \#05 resulted to be one of the hardest when trying to classify per individual 
cameras, although the use of the majority voting strategy on the multiview setup clearly 
contributed to boost the recognition rate (e.g. from $83.1$ to $95$). 
In our opinion, the main difficulty when dealing with this kind of curved trajectories is the 
fragmentation of the person tracks due to partial occlusions (i.e. body parts temporally out of 
camera's field of view), what in turn implies the loss of dense tracks and, therefore, less motion 
features available for characterizing the gait of the subject.

With regard to the number of frames needed to recognize a person with the proposed framework, we can see in Tab.~\ref{tab:resultsMinLen} that with the use of local motion features from 15 consecutive frames, the recognition rate in the multiview setup is nearly perfect ($98.3\%$). Although, in a monocular setup, such configuration reaches a modest $86.3\%$. If we increase the number of used frames up to 50, we obtain an average of $100\%$ in multiview and $96.3\%$ per camera.

If we focus on the results obtained by each of the four used cameras, 
we can see in Tab.~\ref{tab:resultsCam} that camera \textit{`Cam03'} yields a perfect recognition 
rate ($100\%$) on average. Note that for such camera the viewpoint of the person's trajectory is 
nearly profile, thus allowing a suitable computation of point tracks for long time intervals. 
In contrast, the lowest rate is obtained while working 
only with camera \textit{`Cam01'} ($91.7\%$). In our opinion, it is due to the fact that only 
during a short period of time (around $1$ second), the whole body of people is fully 
visible (i.e. neither the head nor the feet are out of the field of view).  

Recall that the trajectories of the densely sampled points, 
used as low level features, are described by using the descriptor DCS (Sec.~\ref{subsec:features}), 
which is compound of four subtypes of features. In Tab.~\ref{tab:resultsFeatDics} we can see the 
effect of removing, at a time, some of the subtypes. The results show that the weakest subtype of 
feature is the \textit{normalized coordinates} as shown in row `PFM-ft1000'. 
In addition to selecting only part of the descriptor, the results of the bottom rows of the table show the effect of dimensionality reduction with PCA. In contrast to the results previously reported in Tab.~\ref{tab:resultsA} where the whole vector of low level features was reduced to a fixed size (e.g. `PCAL100'), in this case, the subtypes are indepedently reduced to a fraction of the original size. For example, `PCALx40' indicates that only the $40\%$ of the dimensions are kept.

\section{Experimental results on MoBo}\label{sec:experesMOBO}
\begin{figure*}[t]
\begin{center}
  \includegraphics[width=0.96 \textwidth]{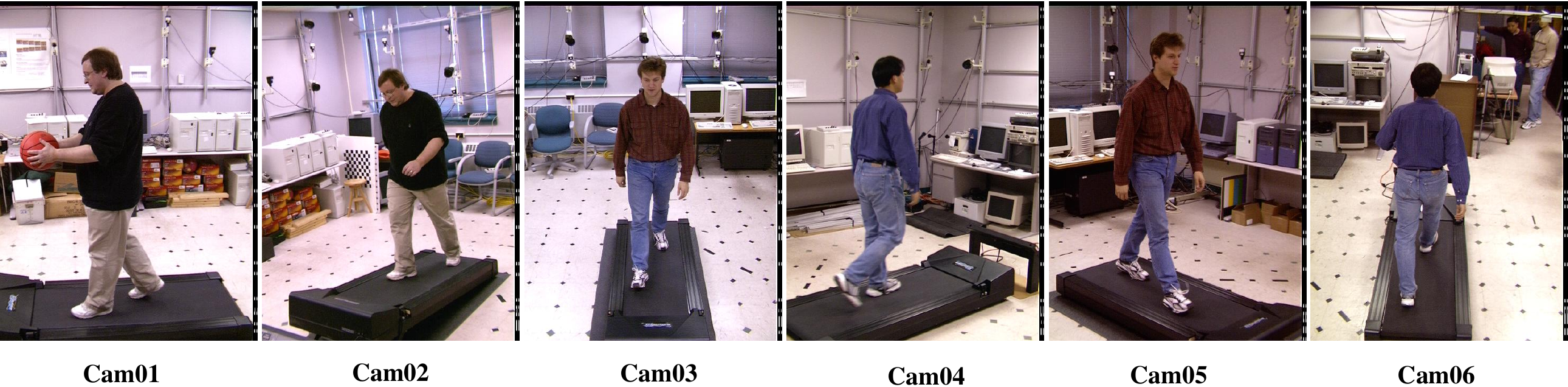}%
\vspace{-0.01cm}	
 \caption{\textbf{MoBo dataset.} Different people recorded from six camera viewpoints walking on a treadmill. Four walking patterns are included in the dataset: ball, incline, slow and fast.}
 \label{fig:MOBOdataset}
\vspace{-0.05cm}
\end{center}
\end{figure*}

The second dataset where we carry out our experiments is the ``CMU Motion of Body'' (MoBo) 
database~\cite{Gross2001mobo}.
MoBo contains video sequences of 25 subjects performing four different walking patterns on a 
treadmill: slow walk, fast walk, incline walk and walking with a ball. It has been recorded from 
six camera viewpoints. Fig.~\ref{fig:MOBOdataset} shows six video frames from the dataset.
Video resolution is $480 \times 640$ pixels.

\subsection{Experimental setup}\label{subsec:expSetupMOBO}
We run here experiments similar to the ones presented above to evaluate further our proposed 
framework for gait recognition. 
Note that, in contrast to AVAMVG dataset where people actually displace along the room, 
in MoBo dataset there is no actual displacement of the person body as people are walking on a treadmill. 
Therefore, there will not be available motion trajectories associated to the body as a whole.
In addition, there is a set of videos where people do not move arms freely as they are holding a ball, 
thus removing the motion pattern associated to arms swing.
In our experiments, we will use only four of the cameras (i.e. \textit{Cam01,Cam02,Cam04,Cam05}). 
As done in the previous section, if multiple cameras
are available during testing,  a majority voting strategy is followed
to deliver a single label per subject.

\noindent \textbf{Experiment A: training on multiple walking patterns.} %
In this experiment we use, at a time, three of the walking patterns for training and the remaining one for testing (i.e. leave-one-out
on the walking patterns). Therefore, we report the average of the four recognition rates. 
The results of this experiment are summarized in Tab.~\ref{tab:resultsFeatDicsMOBO} where each row correspond to a different configuration. We combine different subtypes of DCS features with diverse PCA-based dimensionality reductions and dictionary sizes.

\noindent \textbf{Experiment B: influence of sequence length.} %
The goal of this experiment is to evaluate the influence of the sequence length in the recognition process. For this purpose, at test time, we use only a subsequence of $T$ frames extracted around the middle of the sequence. 
In Tab.~\ref{tab:resultsLenMOBO}, each row corresponds to a different number of frames in the range 
$[5,40]$. The dictionary size has been fixed to either 50 or 100 (column `K') on the configuration 
`PFM+PCALx20+PCAH256-ft0110'.

\noindent \textbf{Experiment C: testing on a single camera.} %
The previous results assume a multicamera enviroment at test time to combine 
information from diverse viewpoints. However, in many real situations, only a viewpoint is available 
at test time. In this experiment we present a breakdown of the recognition rates per camera.
The results of this experiment are summarized in Tab.\ref{tab:resultsCamMOBO}. 
We use here the configuration `PFM+PCALx20+PCAH256-ft0110' with $K=100$ learnt during the previous 
experiment (see Tab.~\ref{tab:resultsLenMOBO}). Each row corresponds to a different camera 
viewpoint.

\noindent \textbf{Experiment D: training on a single walking pattern.} %
In this experiment, we train the models on a single walking pattern, at a time, and then we test on the other patterns.
Tab.~\ref{tab:resultsSingleTrjMOBO} summarizes the results of this experiment. Rows correspond to training walking patterns,
whereas columns correspond to test patterns.
This table allows a comparison with some previously published results on this dataset.

\begin{table*}[t!] %
\renewcommand{\arraystretch}{0.9}
\caption{\textbf{Recognition results on 
MoBo: experiment A.} Each entry contains the percentage of correct 
recognition in the multiview setup and, in parenthesis, the recognition per single view. Each row 
corresponds to a different configuration of the gait descriptor. 
$K$ is the GMM size used for FM. Four camera viewpoints where used during training. Best results are marked in bold.
(See main text for further details.)}
\label{tab:resultsFeatDicsMOBO}
\footnotesize
\centering
\setlength{\tabcolsep}{0.3em} %
\begin{tabular}{l|c| c c c c c}
 \hline
 \textit{Experiment} & $K$ & \textit{Trj=f+i+b} & \textit{Trj=s+f+i} & \textit{Trj=s+f+b} & \textit{Trj=s+i+b} & \textit{Avg}\\
 \hline
  PFM-ft0111       & 100 & 100 (100) & 100 (100) & 100 (100) & 96 (95)  &  99 (98.8)\\
  \hline
  PFM+PCALx40+PCAH256-ft0111  & 50 &  100 (100) & 100 (100) & 100 (100) & 100 (99) & \textbf{100 
(99.8)} \\
  PFM+PCALx40+PCAH064-ft0110  & 50 & 96 (96) & 95.8 (94.8) & 100 (98) & 96 (92)  &  97 (95.2) \\
  PFM+PCALx40+PCAH128-ft0110  & 50 & 100 (100) & 100 (100) & 100 (100) & 100 (99) & \textbf{100 
(99.8)} \\
  PFM+PCALx40+PCAH256-ft0110  & 50 & 100 (100) & 100 (100) & 100 (100) & 100 (99) & \textbf{100 
(99.8)} \\  
  PFM+PCALx20+PCAH064-ft0110  & 50 &  96 (96) & 83.3 (83.3) & 100 (99) & 92 (90) & 92.8 (92.1)\\
  PFM+PCALx20+PCAH128-ft0110  & 50 & 100 (100) & 100 (99) & 100 (100) & 100 (98) &  100 (99.3)\\
  PFM+PCALx20+PCAH256-ft0110  & 50 & 100 (100) & 100 (97.9) & 100 (100) & 100 (98) & 100 (99) \\
  PFM+PCALx20+PCAH128-ft0100  & 100 & \textbf{100 (100)} & \textbf{100 (100)} & \textbf{100 (100)} & 96 (94)&  99 (98.5)\\
  PFM+PCALx20+PCAH128-ft0010  & 100 & 100 (100) & 100 (99) & 100 (99) & \textbf{100 (99)}&  100 (99.3)\\
  PFM+PCALx10+PCAH064-ft0110  & 50 & 96 (96) & 75 (78.1) & 100 (99) & 92 (88) & 90.8 (90.3) \\

  PFM+PCALx10+PCAH256-ft0110  & 100 & 100 (100) & 100 (100) & 100 (100) & 96 (96) &  99 (99)\\
  \hline
  PFM+PCALx20+PCAH128+pyr-ft0110  & 50 & 100 (100) & 95.8 (97.9) & 100 (100) & 100 (99) & 99 (99.2) \\

  PFM+PCALx20+PCAH256+pyr-ft0110  & 100 & 100 (100) & 100 (100) & 100 (100) & 100 (98) &  100 (99.5)\\
 \hline
\end{tabular}
\end{table*}
\subsection{Results}\label{subsec:resultsMOBO}

The results shown in Tab.~\ref{tab:resultsFeatDicsMOBO} correspond to \textit{Experiment A}. Since 
the number of possible combinations of the parameters is so big, we show in the table only a subset 
of the most representative ones. All the configurations (i.e. rows) not containing the suffix `pyr' 
correspond to a single level PFM with two vertical partitions (i.e. concatenation of lower- and 
upper-body FM descriptors). Note that one of the best configurations (in terms of mean accuracy) is 
`PFM+PCALx40+PCAH128-ft0110' with $K=50$, what means that the selected low level motion features 
(DC and CS) have been initially reduced up to the $40\%$ of their 
original dimensionality, and the final PFM descriptor has been reduced to 128 dimensions before the 
classification stage. In such case, we obtain a perfect mean recognition rate in the multiview setup 
and a mean $99.8\%$ of accuracy per viewpoint. This indicates a clear success of the proposed 
descriptor in this dataset with low-dimensional feature vectors.
Focusing on the dimensionality reduction of the final PFM vector, reducing it up to 64 dimensions 
only implies a small decrease in the accuracy ($3\%$ in the multiview case) while obtaining a more 
compact representation of the gait. The most extreme case of compression evaluated in this 
experiment is represented in row `PFM+PCALx10+PCAH064-ft0110' (with $K=50$). However, the accuracy 
worsens less than $10\%$. 
It is worth mentioning that a perfect mean recognition accuracy can be achieved in the multiview 
setup by using only the low level descriptor \textit{CS} (row `PFM+PCALx20+PCAH128-ft0010').
Since so high accuracy is obtained without using more than one level in the PFM, the results 
reported in the bottom rows of Tab.~\ref{tab:resultsFeatDicsMOBO} (including suffix `pyr'), where 
two levels are used, are included just for the completeness of the evaluation.

\begin{table}[th] %
\renewcommand{\arraystretch}{1.1}
\caption{\textbf{Recognition results on MoBo: experiment B.}  Each entry contains the percentage of correct recognition in the multiview setup and, in parenthesis, the recognition per single view. 
Each row corresponds to a particular length of the test sequences. 
$K$ is the GMM size used for FM. %
Best results are marked in bold.
(See main text for further details.)}
\label{tab:resultsLenMOBO}
\centering
\scriptsize
\setlength{\tabcolsep}{0.35em} %
\begin{tabular}{l|c| c c c c c}
 \hline
 \textit{$L$} & $K$ & \textit{Trj=f+i+b} & \textit{Trj=s+f+i} & \textit{Trj=s+f+b} & \textit{Trj=s+i+b} & \textit{Avg}\\
 
  \hline
  05  & 100 & 96 (96) & 91.7 (90.6) & 80 (84.8) & 92 (90) & 89.9 
(90.4) \\ %
  10  & 100 & 100 (100) & 91.7 (91.7) & 92 (96) & 96 (94) & 94.9 (95.4) \\
  15  & 50  & 100 (100) & 95.8 (94.8) & 100 (99) & 96 (94) & 98 (97) \\
  20  & 100 & 100 (100) & 100 (97.9) & 100 (100) & 100 (99) &  \textbf{100 (99.2)}\\
  30  & 100 & 100 (100) & 100 (99) & 100 (100) & 96 (95) & 99 (98.5) \\
  40  & 50 & 100 (100) & 100 (96.9) & 100 (100) & 96 (96) & 99 (98.2)\\
 \hline
\end{tabular}
\end{table}

With regard to \textit{Experiment B}, where the influence of the sequence length in the performance 
is evaluated, the results included in Tab.~\ref{tab:resultsLenMOBO} show that a perfect mean 
recognition rate can be achieved using features of 20 consecutive frames. In addition, by using 
only 10 frames, the accuracy only decreases to approximately $95\%$ of correct recognition.

\begin{table}[th] %
\renewcommand{\arraystretch}{1.0}
\caption{\textbf{Recognition results on MoBo: experiment C.} Each entry contains the percentage of 
correct recognition in the single view setup. The PFM configuration is: PFM+PCALx20+PCAH256-ft0110-len20. Each row corresponds to a different camera. 
$K$ is the GMM size used for FM. (See main text for further details.)}
\label{tab:resultsCamMOBO}
\centering
\scriptsize
\setlength{\tabcolsep}{0.4em} %
\begin{tabular}{l|c| c c c c} %
 \hline
 \textit{Camera} & $K$ & \textit{Trj=f+i+b} & \textit{Trj=s+f+i} & \textit{Trj=s+f+b} & \textit{Trj=s+i+b} \\
 \hline
  \textit{Cam01}       & 100 &  100 & 100 & 100 & 100\\
  \textit{Cam02}       & 100 &  100 & 100 & 95.8 & 95.8\\
  \textit{Cam04}       & 100 &  100 & 100 & 100 & 100\\
  \textit{Cam05}       & 100 &  100 & 96  & 100 & 100 \\
  \hline
\end{tabular}
\vspace{-0.04cm}
\end{table}

The results obtained per camera for the case `PFM+PCALx20+PCAH256-ft0110-len20' (included in 
Tab.~\ref{tab:resultsLenMOBO}) are summarized in Tab.~\ref{tab:resultsCamMOBO}. Note that cameras 
\textit{Cam01} and \textit{Cam04} achieve a mean perfect recognition rate,  what means that not all 
viewpoints are needed for an accurate identification of the individuals with PFM. As previously 
observed in the experiments on `AVAMVG', profile viewpoints favor the recognition process.

\begin{table}[th] %
\renewcommand{\arraystretch}{1.0}
\caption{\textbf{Recognition results on MoBo: experiment D.} Only a single type of trajectory is
used during training. PFM configuration: $K=100$, ft=0110. %
Acronyms: Trn=training pattern, Tst=test pattern, s=slow, f=fast, i=incline, b=ball}
\label{tab:resultsSingleTrjMOBO}
\centering
\footnotesize
\setlength{\tabcolsep}{0.3em} %
\begin{tabular}{l|c c c c}
  \hline
           & Tst=\textit{s}&  Tst=\textit{f} & Tst=\textit{i} & Tst=\textit{b}\\
 \hline Trn=\textit{s} & - & 92 & 100 & 100  \\
  Trn=\textit{f} & 92 & - & 96 & 83.3  \\
  Trn=\textit{i} & 100 & 96 & - & 87.5  \\
  Trn=\textit{b} & 48 & 48 & 44 & -  \\
  \hline
\end{tabular}
\vspace{-0.02cm}
\end{table}
The results shown in Tab.~\ref{tab:resultsSingleTrjMOBO} correspond to \textit{Experiment D}. 
In particular, we have used the following configuration: features DC and CS; 
$K=100$ for the dictionaries; and single level PFM with two vertical partitions. 
No PCA compression has been performed in this experiment.
To increase the 
number of training samples, the video sequences have been split into subsequences of length 100 
frames, with an overlap of 25 frames. In this case, the worst recognition results are obtained when 
the training stage is carried out on the \textit{ball} sequences. This result is reasonable since 
the classifiers have never seen the motion of the arms that is present in the other walking types 
(\textit{s}, \textit{f} and \textit{i}).
To put our results in context, the authors of \cite{collins02fgr} report the results of training on 
\textit{slow walk} and testing on both \textit{fast walk} and \textit{ball}, obtaining $92\%$ and $96\%$ 
of accuracy, respectively. In our case, we obtain the same accuracy on \textit{fast walk} (column 
`Tst=f') and improve on the \textit{ball} case (column `Tst=b'). 
We can also compare with the results published in \cite{liang2006}, 
where in one case they train on \textit{slow walk} and test on \textit{ball}, obtaining $91.7\%$; 
and a second case where they train on \textit{fast walk} and test on \textit{slow walk}, obtaining 
$96\%$ of accuracy. In our case, we obtain $100\%$ and $92\%$, respectively.
Comparing with the results reported in \cite{DasChoudhury20123414}, we outperform the cases of training on \textit{slow walk} and testing on \textit{ball}, and training on \textit{fast walk} and testing on \textit{slow walk}, obtaining $100\%$ and $92\%$ of accuracy, respectively. In the rest of cases, we obtain similar results.
In \cite{Chen2011988}, the authors only perform the experiments in cases of training on \textit{slow walk} and testing on \textit{fast walk} and training on \textit{fast walk} and testing on \textit{slow walk}, obtaining $100\%$ and $92\%$ respectively. In our case, we obtain a $92\%$ in both cases.
In the recent paper \cite{lee2014}, the reported results on this dataset 
only use the combination of training on \textit{fast walk} and testing on \textit{slow walk}, 
achieving a recognition rate of $88\%$, in contrast to our $92\%$.
\section{Experimental results on CASIA}\label{sec:experesCASIA}
The third dataset where we perform our experiments is ``CASIA Gait Dataset''~\cite{casiaDB}, parts B (CASIA-B)~\cite{yu2006casia} 
and C (CASIA-C)~\cite{tan2006casiaC}. 

\begin{figure}[t]
\begin{center}
  \includegraphics[width=0.70 \textwidth]{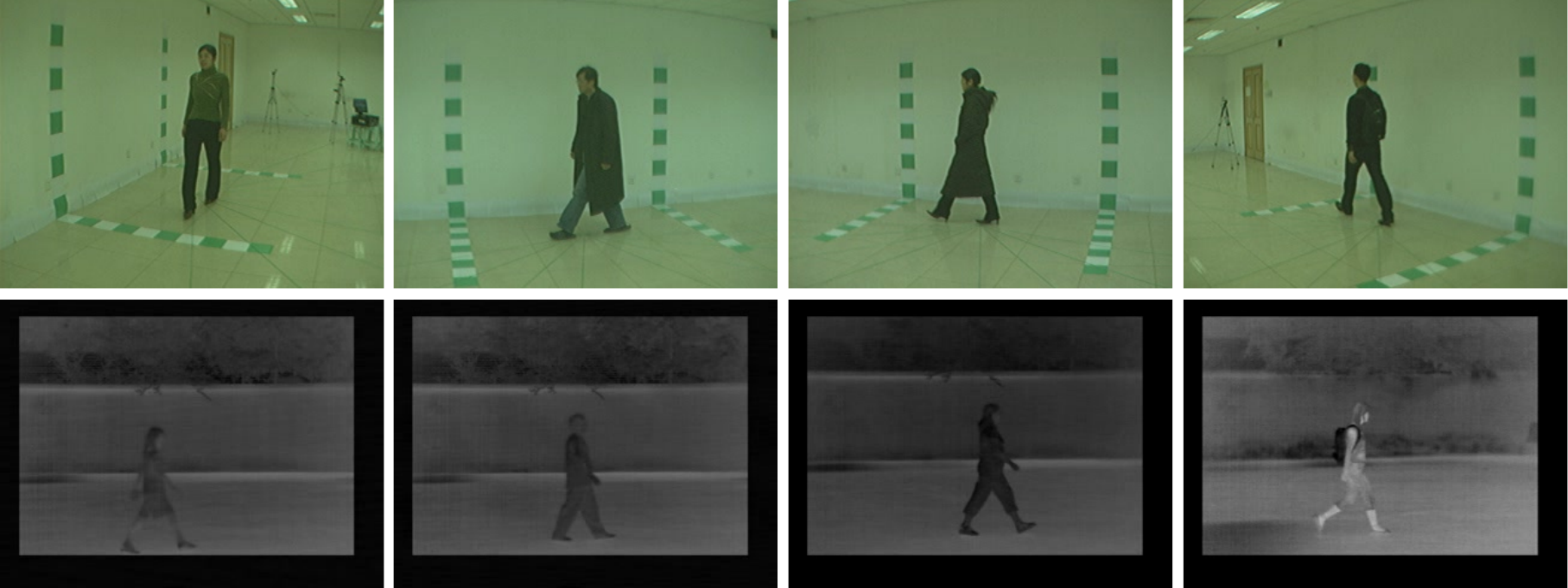}%
\vspace{-0.01cm}
 \caption{\textbf{CASIA dataset.} \textbf{(top)} CASIA-B. People recorded from different camera viewpoints walking indoors. 
 Three situations are included in the dataset: normal walking, walking with a coat and walking with a bag.
 \textbf{(bottom)} CASIA-C. Different people recorded outdoors during night with an infrared camera. Four situations are 
 included in the dataset: normal walking, slow walking, fast walking and walking with a bag.}
 \label{fig:CASIAdataset}
\vspace{-0.05cm}
\end{center}
\end{figure}

In CASIA-B 124 subjects perform walking trajectories in an indoor environment. The action is captured 
from 11 viewpoints. Three situations are considered: normal walk (\textit{nm}), wearing a coat (\textit{cl}), 
and carrying a bag (\textit{bg}).
The first camera viewpoint included in the dataset is $0^o$ and the last one is $180^o$, the intermediate ones
are separated by $18^o$. Some examples can be seen in the top row of Fig.~\ref{fig:CASIAdataset}.
In CASIA-C 153 subjects perform walking trajectories in an outdoor environment during night. The action is captured 
from a single viewpoint with an infrared camera. Four situations are considered: normal walk (\textit{fn}), 
fast walk (\textit{fq}), slow walk (\textit{fs}) and carrying a bag (\textit{fb}).
Some examples can be seen in the bottom row of Fig.~\ref{fig:CASIAdataset}.
In both sets, video resolution is $320 \times 240$ pixels.

\subsection{Experimental setup}\label{subsec:casiaSetup}
We run here experiments similar to the ones presented above to evaluate the performance of our proposed 
framework on videos with low resolution, in comparison to the ones previously used in our experiments. 
In addition, this a more challenging dataset due to the larger amount of people, different scenarios (i.e. wearing
coats and carrying bags), and 
outdoor and night conditions.

We use the full length video sequences to build single layer PFM descriptors, unless otherwise stated.

\begin{table}[th] %
\renewcommand{\arraystretch}{1.1}
\caption{\textbf{Recognition results on CASIA-B: experiment A}. Setup: `nm' scenario, K=100, PCAL=100, PCAH=128. Training on sequences $\{1,2,3,4\}$
and testing on sequences $\{5,6\}$. Each row is a training camera, whereas columns are test cameras.}
\label{tab:casiaBnm}
\centering
\scriptsize
\setlength{\tabcolsep}{0.45em} %
\begin{tabular}{c|ccccccccccc}
\hline 
\textit{Cam} & \textit{0} &\textit{18} & \textit{36} & \textit{54} & \textit{72} & \textit{90} & \textit{108} & \textit{126} & \textit{144} & \textit{162} & \textit{180} \\ 
\hline 
\textit{0} & 100 & 8.1 & 2.4 & 0.8 & 0.8 & 0.8 & 1.2 & 0.8 & 1.2 & 2.0 & 1.2 \\ 

\textit{18} & 17.3 & 100 & 95.2 & 23.8 & 5.9 & 4.4 & 2.4 & 2.0 & 3.2 & 2.0 & 0.8 \\ 

\textit{36} & 2.0 & 94.0 & 100 & 99.6 & 56.5 & 18.1 & 21.4 & 31.0 & 23.8 & 8.1 & 1.6 \\ 
 
\textit{54} & 1.6 & 37.9 & 98.4 & 100 & 99 & 67.7 & 70.2 & 73.8 & 37.9 & 5.6 & 1.2 \\ 
 
\textit{72} & 1.2 & 3.6 & 23.0 & 79.4 & 100 & 80.2 & 77.4 & 62.9 & 16.9 & 4.8 & 2.0 \\ 
 
\textit{90} & 1.6 & 4.0 & 10.1 & 66.1 & 99.5 & 99.6 & 94.4 & 33.1 & 7.7 & 3.6 & 0.8 \\ 
 
\textit{108} & 0.8 & 7.3 & 16.9 & 60.9 & 86.5 & 97.2 & 99.6 & 97.6 & 55.2 & 10.1 & 1.6 \\ 
 
\textit{126} & 0.4 & 8.1 & 37.9 & 68.5 & 76.5 & 76.6 & 95.6 & 98.8 & 96.4 & 19.4 & 1.2 \\ 
 
\textit{144} & 1.6 & 12.1 & 32.7 & 29.8 & 10.5 & 12.5 & 53.2 & 95.6 & 99.2 & 12.5 & 2.4 \\ 
 
\textit{162} & 4.8 & 7.3 & 2.8 & 2.4 & 3.0 & 2.4 & 2.4 & 2.8 & 15.7 & 100 & 11.7 \\ 

\textit{180} & 6.0 & 1.2 & 1.2 & 0.8 & 1.6 & 1.2 & 1.2 & 1.2 & 1.6 & 3.6 & 100 \\

\hline 
\end{tabular}
\vspace{-0.03cm}
\end{table}
\noindent \textbf{Experiment A: training on a single camera and testing on different cameras.} %
The goal of this experiment is to evaluate the capacity of generalization 
of the proposed system to changes in camera viewpoint. 
This experiment is carried out on CASIA-B. We use a model trained on sequences 1 to 4 of the `nm' scenario
to classify sequences 5 and 6 of the same scenario.
The recognition percentages for this experiment are summarized in table Tab.~\ref{tab:casiaBnm}.
Each row correspond to a training camera viewpoint, whereas the columns correspond to test camera viewpoints.

\begin{table}[th] %
\renewcommand{\arraystretch}{1.1}
\caption{\textbf{Recognition results on CASIA-B: experiment B-\textit{cl}}. Setup: K=100, PCAL=150, PCAH=256. Training on `nm' sequences $\{1,2,3,4\}$
and testing on `cl' sequences $\{1,2\}$. Each row is a training camera, whereas columns are test cameras.}
\label{tab:casiaBcl}
\centering
\scriptsize
\setlength{\tabcolsep}{0.45em} %
\begin{tabular}{c|ccccccccccc}
\hline 
\textit{Cam} & \textit{0} & \textit{18} & \textit{36} & \textit{54} & \textit{72} & \textit{90} & \textit{108} & \textit{126} & \textit{144} & \textit{162} & \textit{180}\\ 
\hline 
\textit{0} & 96.8 & 4.5 & 2.4 & 1.2 & 1.2 & 0.8 & 1.2 & 1.6 & 1.2 & 1.2 & 0.8 \\

\textit{18} & 6.9 & 92.3 & 45.6 & 7.7 & 3.2 & 1.6 & 1.6 & 1.2 & 1.2 & 1.2 & 0.8 \\

\textit{36} & 1.6 & 54.5 & 90.3 & 68.1 & 29.8 & 13.3 & 12.9 & 12.9 & 6.9 & 3.6 & 0.8 \\ 

\textit{54} & 0.8 & 17.9 & 63.7 & 89.5 & 76.2 & 49.2 & 37.9 & 28.6 & 10.5 & 6.0 & 0.8 \\ 

\textit{72} & 0.4 & 0.4 & 14.9 & 52.4 & 70.6 & 59.3 & 41.1 & 22.2 & 7.3 & 2.0 & 0.8 \\ 

\textit{90} & 2.0 & 3.7 & 5.6 & 17.3 & 63.7 & 82.3 & 41.9 & 14.1 & 4.8 & 5.6 & 0.8 \\ 

\textit{108} & 1.6 & 3.7 & 5.6 & 18.1 & 41.1 & 63.3 & 83.1 & 55.2 & 13.3 & 4.0 & 0.8 \\ 

\textit{126} & 0.8 & 4.5 & 8.1 & 27.8 & 34.3 & 41.1 & 66.9 & 81.5 & 54.8 & 8.9 & 0.8 \\ 

\textit{144} & 0.8 & 4.9 & 12.5 & 12.5 & 11.7 & 10.9 & 28.2 & 58.9 & 87.9 & 1.6 & 1.6 \\ 

\textit{162} & 2.4 & 2.0 & 3.2 & 2.4 & 2.4 & 2.4 & 2.4 & 2.8 & 2.4 & 91.9 & 2.8 \\ 

\textit{180} & 4.4 & 0.8 & 0.8 & 0.8 & 0.8 & 0.8 & 0.8 & 0.8 & 0.8 & 1.6 & 87.5 \\  
\hline 
\end{tabular}
\vspace{-0.03cm}
\end{table}
\begin{table}[th] %
\renewcommand{\arraystretch}{1.1}
\caption{\textbf{Recognition results on CASIA-B: experiment B-\textit{bg}}. Setup: K=150, PCAL=150, PCAH=256. Training on `nm' sequences $\{1,2,3,4\}$
and testing on `bg' sequences $\{1,2\}$. Each row is a training camera, whereas columns are test cameras.}
\label{tab:casiaBbg}
\centering
\scriptsize
\setlength{\tabcolsep}{0.45em} %
\begin{tabular}{c|ccccccccccc}
\hline 
\textit{Cam} & \textit{0} & \textit{18} & \textit{36} & \textit{54} & \textit{72} & \textit{90} & \textit{108} & \textit{126} & \textit{144} & \textit{162} & \textit{180}\\ 
\hline 
\textit{0} & 99.6 & 5.7 & 2.4 & 2.4 & 2.4 & 2.4 & 1.6 & 1.6 & 1.6 & 2.0 & 2.0 \\

\textit{18} & 11.7 & 99.2 & 83.9 & 16.9 & 4.4 & 2.4 & 3.6 & 1.6 & 1.2 & 2.0 & 0.8 \\

\textit{36} & 2.8 & 86.9 & 100 & 96 & 50.4 & 25.0 & 22.2 & 28.2 & 21.8 & 7.7 & 1.6\\ 
 
\textit{54} & 0.8 & 30.3 & 92.3 & 100 & 95.2 & 71.8 & 61.3 & 60.1 & 27.8 & 3.2 & 0.8\\ 
 
\textit{72} & 2.8 & 0.8 & 27.4 & 74.6 & 80.6 & 79.4 & 72.6 & 50.4 & 12.1 & 8.1 & 1.6\\ 
 
\textit{90} & 1.6 & 4.1 & 6.9 & 39.9 & 97.2 & 100 & 74.2 & 22.2 & 8.1 & 4.4 & 2.4\\ 
 
\textit{108} & 1.2 & 4.5 & 10.9 & 37.9 & 63.7 & 87.1 & 99.2 & 89.9 & 37.1 & 6.9 & 2.8 \\ 
 
\textit{126} & 0.8 & 7.8 & 24.6 & 51.2 & 62.5 & 69.8 & 94.4 & 98.8 & 91.1 & 19.0 & 0.8\\ 
 
\textit{144} & 0.8 & 10.2 & 30.2 & 27.8 & 19.0 & 18.5 & 42.3 & 87.9 & 98.8 & 7.7 & 2.0 \\ 

\textit{162} & 3.6 & 8.6 & 3.2 & 2.4 & 2.4 & 2.4 & 2.4 & 4.0 & 4.8 & 99.6 & 8.1 \\ 

\textit{180} & 7.7 & 2.9 & 0.8 & 1.2 & 0.8 & 0.8 & 0.8 & 1.2 & 2.4 & 4.0 & 99.6 \\
\hline 
\end{tabular}
\vspace{-0.03cm}
\end{table}
\begin{table}[th] %
\renewcommand{\arraystretch}{1.0}
\caption{\textbf{State-of-the-art on CASIA-B, camera $90^o$}. Percentage of correct recognition for several methods on camera $90^o$. Acronyms: `\#subjs' number of subjects used for test; `\#train' number of sequences per person used for training; `\#test' number of sequences per person used for test. Best results are marked in bold. }
\label{tab:casiaBcomp90}
\centering
\footnotesize
\setlength{\tabcolsep}{0.4em} %
\begin{tabular}{c|ccc|ccc|c}
\hline 
\textit{Method} & \#subjs & \#train & \#test & nm & bg & cl & \textit{Avg} \\ 
\hline 
AEI+2DLPP~\cite{AEI}& 124 & 3 & \specialcell{3-nm\\2-bg-cl} & 98.4 & 91.9 & 72.2 & 87.5 \\
GEI~\cite{yu2006casia} & 124 & 4 & 2 & 97.6 & 52.0 & 32.7 &  67.8 \\
iHMM~\cite{IHMM} & 84 & 5 & 1 & 94.0 & 45.2 & 42.9 &  60.7 \\
CGI~\cite{CGI} & 124 & 1 & 1 & 88.1 & 43.7 & 43.0 & 58.3 \\
VI-MGR~\cite{DasChoudhury2015798} & 124 & 4 & 2 & 100 & 89.0 & 76.0 &  88.3 \\
SDL~\cite{SDL}& 124 & 3 & \specialcell{3-nm\\2-bg-cl} & 98.4 & 93.5 & \textbf{90.3} & 94.1 \\
\hline
PFM (this paper) & 124 & 4 & 2 &  \textbf{100} & \textbf{100} & 85.5 & \textbf{95.2} \\
\hline 
\end{tabular}
\vspace{-0.03cm}
\end{table}
\noindent \textbf{Experiment B: robustness to clothing and carrying objects.} %
The goal of this experiment is to evaluate the robustness of our proposed system to changes in shape due to 
changing clothing and carrying bags. 
This experiment is carried out on CASIA-B. We use a model trained on sequences 1 to 4 of the `nm' scenario
to classify sequences 1 and 2 of scenarios `cl' (wearing a coat) and `bg' (carrying a bag).
The recognition percentages for this experiment are summarized in tables Tab.~\ref{tab:casiaBcl} and Tab.~\ref{tab:casiaBbg}.
Each row correspond to a training camera viewpoint, whereas the columns correspond to test camera viewpoints.

\begin{table*}[ht] %
\renewcommand{\arraystretch}{1.1}
\caption{\textbf{Recognition results on CASIA-B: experiment C}. Setup: PCAL = 150 and PCAH = 256, unless otherwise stated. Training on `nm' sequences $\{1,2,3,4\}$
and testing on remaining sequences. Each row is a set of training camera viewpoints, whereas columns are test cameras. Sufix `sil' indicates that binary silhouettes where used to define people location instead of the person detector. Values in italics are included just for reference, but they are not used in the computation of `Avg'.}
\label{tab:casiaBmvnm}
\centering
\scriptsize
\setlength{\tabcolsep}{0.3em} %
\begin{tabular}{c|c|ccccccccccc|c}
\hline 
\textit{Test} & \textit{Training cams} & \textit{0} & \textit{18} & \textit{36} & \textit{54} & \textit{72} & \textit{90} & \textit{108} & \textit{126} & \textit{144} & \textit{162} & \textit{180} & \textit{Avg} \\ 
\hline 
\multirow{2}{*}{\textit{nm}} & 18, 54, 90, 126, 162 (PCA100-K150) & \textit{8.5} & 100 & 100 & 100 & 100 & 100 & 100 & 100 & 100 & 100 & \textit{8.1 }&  100 \\ 

 &18,36,54,72,90,108,126,144,162 (PCA150-K150) & \textit{4}&  100 & 100 & 100 & 100 & 100 & 100 & 100 & 100 & 100 & \textit{2.8} &  100 \\ 
& all (K100) & 98.8 & 99.6 & 100.0 & 100.0 & 100.0 & 100.0 & 100.0 & 100.0 & 100.0 & 100.0 & 98.8 & 99.7 \\
\hline
\hline
\multirow{2}{*}{\textit{bg}} & 18, 54, 90, 126, 162 (K150) & \textit{5.2} & 98.4 & 96.0 & 98.0 & 97.2 & 98.0 & 96.0 & 98.8 & 92.7 & 98.4 & \textit{5.2} & 97.1 \\ 

&18, 36, 54, 72, 90, 108, 126, 144, 162 (K150) & \textit{5.2} & 97.1 & 97.6 & 98.4 & 100 & 98.8 & 98.0 & 97.6 & 97.6 & 97.2 & \textit{4} & 98.0 \\ 

& all (K100) & 89.5 & 94.3 & 94.4 & 95.2 & 96.8 & 97.6 & 96.0 & 95.2 & 93.1 & 92.3 & 83.9 & 93.5 \\
\hline 
\hline
\multirow{4}{*}{\textit{cl}} & 18, 54, 90, 126, 162 (K150) & \textit{2} & 73.2 & 65.7 & 72.6 & 70.6 & 76.2 & 68.1 & 72.6 & 53.6 & 73.4 &\textit{2}&  69.6 \\%72.7 \\ 

&18, 36, 54, 72, 90, 108, 126, 144, 162 (K150) & \textit{2} & 66.3 & 72.2 & 72.2 & 75.0 & 74.6 & 73.8 & 74.2 & 71.4 & 65.7 & \textit{1.6} & 71.7 \\%74.5 \\ 
&18, 36, 54, 72, 90, 108, 126, 144, 162 (K100-pyr)&\textit{1.6} & 70.3 & 73.0 & 73.8 & 81.0 & 77.8 & 77.0 & 78.6 & 74.6 & 66.9 & \textit{1.6} & 74.8 \\%74.5 \\ 
&18, 36, 54, 72, 90, 108, 126, 144, 162 (K100-sil) & \textit{3.2} & 75.4 & 76.4 &74.4 &79.0& 81.9& 80.5& 77.3& 75.3& 65.7 & \textit{4.4} & 76.2 \\% GMM silhouettes
& all (K100) & 53.2 & 70.7 & 74.2 & 69.0 & 74.2 & 73.0 & 69.8 & 71.0 & 71.8 & 61.3 & 43.1 & 66.5 \\ 
\hline
 
\end{tabular}
\vspace{-0.03cm}
\end{table*}
\begin{table*}[ht] %
\renewcommand{\arraystretch}{1.1}
\caption{\textbf{Recognition results on CASIA-B: experiment C}. At test time, a subject is viewed from several cameras, thus the opinion of each viewpoint is combined
to decide the identity of the target subject. Column \textit{`Acc'} contains the percentage of correct recognition 
in the multiview setup. In parenthesis, monocular accuracy.  PCAH=256 is used in all cases for the final PFM descriptor. (See main text for further details.)
}
\label{tab:casiaBmvtst}
\centering
\scriptsize
\setlength{\tabcolsep}{0.4em} %
\begin{tabular}{c|c|c|c|c|c}
\hline 
 \textit{Exper.} & \textit{PCAL} & $K$ & \textit{Training cams} & \textit{Test cams} & \textit{Acc}\\ 
\hline 
\textit{nm-nm} & 100 & 150 & 18,54,90,126,162 & 18,36,72,108,126,162 & 100 (100)\\
\textit{nm-bg} & 150 & 100 & 18,36,54,72,90,108,126,144,162 & 18,36,72,108,126,162 & 100 (98.2)\\
\textit{nm-cl} & 150 & 100 & 18,36,54,72,90,108,126,144,162 & 18,36,72,108,126,162 & 83.1 (75)\\
\hline 
\end{tabular}
\vspace{-0.03cm}
\end{table*}
\begin{table*}[th] %
\renewcommand{\arraystretch}{1.1}
\caption{\textbf{State-of-the-art on CASIA-B, multiview training}. Percentage of correct recognition when multiple viewpoints are combined during training. Each column corresponds to a test camera viewpoint of scenario `nm'. Best results are marked in bold. }
\label{tab:casiaBcompMV}
\centering
\scriptsize
\setlength{\tabcolsep}{0.25em} %
\begin{tabular}{c|ccc|ccccccccccc|c}
\hline 
\textit{Method} & \#subjects & \#train-seqs & \#test-seqs & \textit{0} & \textit{18} & \textit{36} & \textit{54} & \textit{72} & \textit{90} & \textit{108} & \textit{126} & \textit{144} & \textit{162} & \textit{180} & \textit{Avg} \\ 
\hline 
IF+iHMM~\cite{IHMM} & 84 & 5 & 1 & 98.8 & 98.8 & 94.0 & 94.0 & 92.9 & 94.0 & 94.0 & 95.2 & 97.6 & 98.8 & \textbf{100} & 96.2 \\ 
GEI+PCA+LDA~\cite{IHMM}~\cite{HUMANID} & 84 & 5 & 1 & 96.4 & 92.9 & 96.4 & 91.7 & 90.5 & 90.5 & 92.9 & 90.5 & 90.5 & 95.2 & 95.2 & 93.0 \\ 
\hline 
PFM (this paper) & 124 & 4 & 2 & \textbf{98.8} & \textbf{99.6} & \textbf{100} & \textbf{100} & \textbf{100} & \textbf{100} & \textbf{100} & \textbf{100} & \textbf{100} & \textbf{100} & 98.8 & \textbf{99.7}\\ 
\hline 
\end{tabular} 
\vspace{-0.04cm}
\end{table*}
\noindent \textbf{Experiment C: training and testing on multiple cameras.} %
CASIA-B

In this experiment, firstly, we use several camera viewpoints to train a single model, and we test on
different viewpoints independently. The results of this experiment are presented in Tab.~\ref{tab:casiaBmvnm}.
We show in the rows different number of cameras used during training. In the case we use all cameras 
for training (from $0^o$ to $180^o$) the mean recognition accuracy over all the test camera viewpoints is $99.7\%$
for the `nm' scenario.

We also evaluate in this experiment the behaviour of the system if multiple viewpoints are available at test time.
Thus, combining the opinion of the individual viewpoints. 
The results are summarized in Tab.~\ref{tab:casiaBmvtst}, where each row corresponds to a different test scenario (e.g. `nm-bg' means training on `nm' sequences and testing on `bg' sequences). Column `Acc' shows the percentage of recognition achieved when the individual opinions of the test cameras (column `Test cams') are combined by majority voting.

\begin{table}[t] %
\renewcommand{\arraystretch}{0.8}
\caption{\textbf{State-of-the-art on CASIA-C}. Percentage of correct recognition on CASIA-C for diverse methods. Each column corresponds to a different scenario. Best results are marked in bold. (See main text for further details).}
\label{tab:casiaCcomp}
\centering
\footnotesize
\setlength{\tabcolsep}{0.3em} %
\begin{tabular}{c|c|cccc}
\hline 
\textit{Method} & \#subjects & \textit{fn} & \textit{fs} & \textit{fq} & \textit{fb} \\ 
\hline 
Gait Curves~\cite{DeCann2010} & 153 & 91.0 & 65.4 & 69.9 & 25.5 \\  
NDDP~\cite{NDDP} & 153 & 98.0 & 84.0 & 84.0 & 16.0 \\ 
ODP~\cite{ODP} & 153 & 98.0 & 80.0 & 80.0 & 16.0 \\ 
WPSR~\cite{WPSR} & 153 & 93.0 & 83.0 & 85.0 & 20.0 \\ 
HTI~\cite{tan2006casiaC} & 46 & 94.0 & 85.0 & 88.0 & 51.0 \\ 
HDP~\cite{HDP} & 153 & 98.0 & 84.0 & 88.0 & 36.0 \\ 
AEI~\cite{AEI} & 153 & 88.9 & 89.2 & 90.2 & 79.7 \\ 
Pseudoshape~\cite{Pseudoshape} & 153 & 98.4 & 91.3 & 93.7 & 24.7 \\ 
WBP~\cite{WBP} & 153 & 99.0 & 86.4 & 89.6 & 80.1 \\ 
HSC~\cite{HSC} & 50 & 98.0 & 92.0 & 92.0 & - \\ 
DCM~\cite{DCM} & 120 & 97.0 & 92.0 & 93.0 & - \\ 
RSM~\cite{RSM} & 153 & 100 & \textbf{99.7} & 99.6 & 96.2 \\ 
\hline 
PFM (this paper) & 153 & \textbf{100} & 98.7 & \textbf{100} & \textbf{99.3} \\ 
\hline 
\end{tabular}
\vspace{-0.04cm}
\end{table}
\noindent \textbf{Experiment D: gait recognition during night.} %
The goal of this experiment is to evaluate the performance of our proposed PFM descriptor on infrared images
taken outdoors.
For this purpose, we use CASIA-C dataset.
Since the previously used person detector (Sec.~\ref{subsec:detection}) showed a poor performance on the infrared images of CASIA-C, 
we carried out background segmentation to define the bounding-box of the persons in these
sequences. For that purpose, we learnt a Gaussian Mixture Model from 40 video frames. We use the implementation
of \cite{kaewtrakulpong2002} included in Matlab. For each video frame, a bounding-box is fitted to the 
obtained foreground pixels,
ensuring a fixed aspect ratio of $1:3$. The remaining stages remain as explained in Sec.~\ref{subsec:detection}.

The bottom row of Tab.~\ref{tab:casiaCcomp} shows the recognition percentages
achieved by our system. For each subject, two sequences from subset `fn' (normal walk) are used for training,
and the remaining sequences are used for testing. Note that, for example, row \textit{RSM}~\cite{RSM} used three
sequences per subject during training, instead of the two we use. 
In our case, the PFM setup for column `fn' is PCAL100+PCAH128+K50; for column `fs' is PCAL150+PCAH256+K150; 
for column `fq' is PCAL100+PCAH256+K50; and, for column `fb' is PCAL150+PCAH128+K50. For the case `fs',
dictionary size $K$ can be reduced to $100$ when using two layers in PFM.

\noindent \textbf{Experiment E: effect of person detection on the system performance.} %
The goal of this experiment is to evaluate the impact of the person detection module (Sec.~\ref{subsec:detection}) 
on the final performance of our system. 
In particular, instead of using our detection module, we use  
binary silhouettes, obtained through a GMM-based background segmentation
(as done with CASIA-C), to filter out the initially estimated dense tracks.
We have tried this for the hardest scenario of CASIA-B, wearing coats (`cl'). The results are included as a row of the Tab.~\ref{tab:casiaBmvnm}. It is indicated with the keyword `sil'.
 Although it is not included in the table, we also tried for this experiment the binary silhouettes 
provided by the authors of the dataset, but we obtained lower results -- we realised that some silhouettes were missing. 
Additionally, in Tab.~\ref{tab:casiaBcomp90}, our accuracy $85.5\%$ for `cl' increased up to $88.3\%$ when using background segmentation.

\subsection{Results}\label{subsec:casiaRes}

The results shown in table Tab.~\ref{tab:casiaBnm} correspond to \textit{experiment A}. 
We can see that in most cases, by training and testing on the same camera, we can obtain an almost perfect identification of the subjects.
This value decreases when testing on different cameras. Although, for similar viewpoints, in some cases we can find drops in accuracy lower than $5\%$ (e.g. $90^o$ vs. $72^o$ and $108^o$). %

If the previously trained system is tested with people wearing coats (`cl'), we can see in Tab.~\ref{tab:casiaBcl} that testing on the same camera where it was trained achieves an average accuracy of $86.7\%$ (i.e. mean of the diagonal). 
We can see in Fig.~\ref{fig:CASIAdataset} (top row) some examples of people wearing a coat that clearly show the difficulty of this scenario, where legs of some people are occluded up to the ankle (i.e. 17 people wear that kind of coats, and seven hold their hands inside the pockets).
In the case of people wearing bags (`bg'), the average accuracy obtained from the diagonal of Tab.~\ref{tab:casiaBbg} is $97.8\%$.
To put our results in context with other works, Tab.~\ref{tab:casiaBcomp90} contains the results of training and testing on only camera $90^o$ for the three scenarios. Bottom row (PFM) shows our best results for each scenario, along with the average performance. Note how our approach improves on the state-of-the-art average from $94.1\%$ to $95.2\%$.

In the previous experiments, an independent classifier was trained per viewpoint. In contrast, Tab.~\ref{tab:casiaBmvnm} shows that a single classifier can be trained including several viewpoints (\textit{Experiment C}). Column `Training cams' indicates the cameras used during training, whereas the subsequent columns indicate the test cameras. Note that for the scenarios `nm' and `bg' the recognition rate is nearly perfect for all the test viewpoints. In all cases, the lowest scores are located on frontal and back viewpoints (i.e. $0^o$ and $180^o$) -- where dense tracklets present very small displacements and, therefore, are less discriminative -- decreasing the average performance (column `Avg'). 
In the case of scenario `cl', the average accuracy increases when using a two-levels PFM, as shown in row `K100-pyr'.
These results are comparable to state-of-the-art ones, as shown in Tab.~\ref{tab:casiaBcompMV}, where our approach improves on the best known result, up to our knowledge, from $96.2\%$ to $99.7\%$, using even less training sequences (four per person instead of five).
The results of the second part of \textit{Experiment C} can be seen in Tab.~\ref{tab:casiaBmvtst}. 
We can see that the majority voting strategy on the test cameras allows to achieve higher recognition rates, as previously shown on the other datasets. For example, in the case `cl', if we consider that each viewpoint is an independent video sequence (monocular case), we obtain an accuracy of $75\%$, but this value grows up to $83.1\%$ in the multiview setup.

With regard to \textit{Experiment E}, in the case of the scenario `cl' (see row with `K100-sil' of
Tab.~\ref{tab:casiaBmvnm}) the average recognition improves from $71.7\%$ to $76.2\%$ (same PFM setup but $K=100$).
What, in our opinion, indicates that our person detector has a good behaviour in general, but should be improved 
for the cases where people wear clothing that deforms the expected shape of a person (from the point of view of the full-body person detector).

Finally, we focus on CASIA-C. The results of \textit{Experiment D}, summarized in row `PFM' of Tab.~\ref{tab:casiaCcomp}, indicate that our proposed descriptor is suitable for outdoors and infrared images, achieving perfect recognition results on two out of four situations, despite the difficulty of this kind of images as can be seen in the bottom row of Fig.~\ref{fig:CASIAdataset}. 
If we compare with previous state-of-the-art approaches\footnote{%
Rows of Tab.~\ref{tab:casiaCcomp} have been imported from the publication \cite{RSM}.
}, %
we can see that our system establishes the new state-of-the-art on two scenarios of this dataset, showing similar results on the other two.

\begin{figure}[t]
\begin{center}
  \includegraphics[width=0.49\textwidth]{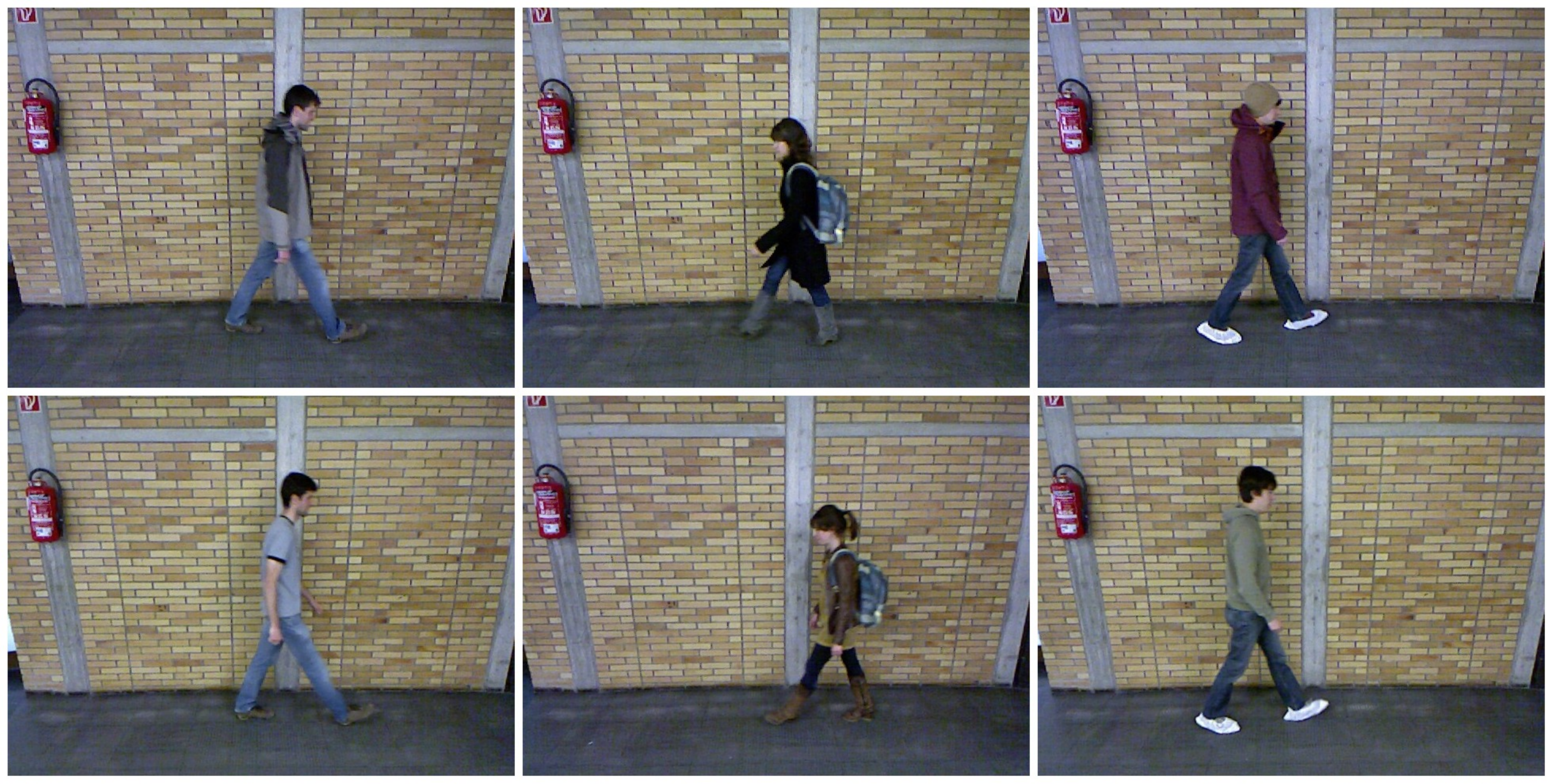}%
\vspace{-0.01cm}	
 \caption{\textbf{TUM GAID dataset.} People recorded from the same viewpoint walking indoors in two seasons. Three situations are included in the dataset: normal walking, walking with a bag and walking with coating shoes. 
 \textbf{(top)} First session (cold day). 
 \textbf{(bottom)} Second session (warm day).}
 \label{fig:TUMdataset}
\vspace{-0.05cm}
\end{center}
\end{figure}
\section{Experimental results on TUM GAID}\label{sec:experesTUM}

The fourth dataset where we perform our experiments is ``TUM Gait from Audio, Image and Depth (GAID) database''~\cite{tumDB}. 

In TUM GAID 305 subjects perform two walking trajectories in an indoor environment. %
The first trajectory is performed from left to right and the second one from right to left. Therefore, both sides of the subjects are recorded. Two recording sessions were performed, one in January, where subjects wear heavy jackets and mostly winter boots, and the second in April, were subjects wear different clothes. Some examples can be seen in the Fig.~\ref{fig:TUMdataset}. 
Top row shows three subjects recorded in the first session. Each one is recorded in a different walking condition (normal walk, carrying a backpack and wearing coating shoes). Bottom row shows the same three subjects recorded in the second session under the same walking conditions as in the first session.

Hereinafter the following nomenclature is used to refer every of the four walking conditions considered: normal walk (\textit{N}), carrying a backpack of approximately 5 kg (\textit{B}), wearing coating shoes (\textit{S}), as used in clean rooms for hygiene conditions, 
and elapsed time (\textit{TN-TB-TS}).

Each subject of the dataset is composed of: six sequences of normal walking (\textit{N1, N2, N3, N4, N5, N6}), two sequences carrying a bag (\textit{B1, B2}) and two sequences wearing coating shoes (\textit{S1, S2}). In addition, 32 subjects were recorded in both sessions  (i.e. January and April) so they have 10 additional sequences (\textit{TN1, TN2, TN3, TN4, TN5, TN6, TB1, TB2, TS1, TS2}). %

The action is captured by a Microsoft Kinect sensor which provides a video stream, a depth stream and four-channel audio. Video and depth are recorded at a resolution of $640 \times 480$ pixels with a frame rate of approximately 30 fps. The four-channel audio is sampled with 24 bits at 16 kHz. In our experiments, we will only use the video stream.

In~\cite{tumDB}, Hofmann et al. designed the recommended experiments that should be performed in the database. For that purpose, they split the database in 3 partitions: 100 subjects for training and building models, 50 subjects for validation and 155 subjects for testing. Finally, a set of experiments (Sec.~\ref{subsec:expSetupTUM}) are proposed for validating the robustness of the algorithms against different factors.

\subsection{Experimental setup}\label{subsec:expSetupTUM}

We run here the experiments proposed by the authors of the dataset at~\cite{tumDB}. Note that we only perform the experimentation in the RGB video stream, leaving depth and audio streams for future improvements of our algorithm.

In our experiments, we use the training and validation sets combined as in~\cite{tumDB} for building the dictionary. The test set is used for training and testing the person-specific classifiers.

In order to increase the amount of training samples, we generate their \textit{mirror} sequences, thus, doubling the amount of samples available during learning. 

\noindent \textbf{Experiment A: identification over different walking conditions} %

In this experiment we use the four first sequences of normal walking (\textit{N1, N2, N3 N4})  for training and the sequences \textit{N5, N6, B1, B2, S1, S2} of normal walking, bag and coating shoes respectively, for testing. 
Therefore, we build the classifier using normal walking, whereas, test is performed over different walking conditions (normal, bag and coating shoes).
 
The results of this experiment are summarized in Tab.~\ref{tab:resultsTUMA} where each row correspond to a different configuration.

\begin{table}[t] %
\renewcommand{\arraystretch}{0.9}
\caption{\textbf{Recognition results on TUM GAID: experiment A.} Each entry contains the percentage of correct 
recognition. Each row corresponds to a different configuration of the gait descriptor. 
$K$ is the GMM size used for FM. Best results are marked in bold.
(See main text for further details.)}
\label{tab:resultsTUMA}
\centering
\footnotesize
\setlength{\tabcolsep}{0.5em} %
\begin{tabular}{l|c| c c c c}
 \hline
 \textit{Experiment} & $K$ & \textit{N} & \textit{B} & \textit{S} & \textit{Avg}\\
  \hline
  PFM+PCAL100+PCAH256  & 200 &  \textbf{99.7} & 98.1 & 96.1 & 98.0 \\
  PFM+PCAL100+PCAH256  & 400 &  98.7 & 96.1 & 96.5 & 97.1 \\
  PFM+PCAL100+PCAH256  & 600 &  99.4 & \textbf{99.0} & \textbf{99.0} & \textbf{99.1} \\
  PFM+PCAL150+PCAH256  & 200 &  99.4 & 97.4 & 98.1 & 98.3 \\
  PFM+PCAL150+PCAH256  & 400 &  99.7 & 98.1 & 97.7 & 98.5 \\
  PFM+PCAL150+PCAH256  & 600 &  99.4 & 98.4 & 98.7 & 98.8 \\
 \hline
\end{tabular}
\end{table}

\noindent \textbf{Experiment B: identification over temporal test set.} %
The goal of this experiment is to evaluate the influence of recording in different seasons with changes in illumination, clothes, etc.

In this experiment, we use for training the first four sequences of normal walking recorded during the first session (\textit{N1, N2, N3 N4}),
and, for testing, sequences \textit{TN1, TN2, TB1, TB2, TS1, TS2} of normal walking, bag and coating shoes, respectively, recorded during the second session.

The results of this experiment are summarized in Tab.~\ref{tab:resultsTUMB}, where each row corresponds to a different configuration. For this experiment we have used a temporal partitioning to increase the number of samples due to the low number of subjects available for training (16 subjects). All temporal partitions have an overlap of 10 frames between partitions. Further details are presented below.
\begin{table}[t] %
\renewcommand{\arraystretch}{0.9}
\caption{\textbf{Recognition results on 
TUM GAID: experiment B.} Each entry contains the percentage of correct 
recognition. Each row corresponds to a different configuration of the gait descriptor. 
$K$ is the GMM size used for FM. `len' indicates the number of frames used in temporal partition. Best results are marked in bold.
(See main text for further details.)}
\label{tab:resultsTUMB}
\centering
\footnotesize
\setlength{\tabcolsep}{0.35em} %
\begin{tabular}{l|c| c c c c}
 \hline
 \textit{Experiment} & $K$ & \textit{TN} & \textit{TB} & \textit{TS} & \textit{Avg}\\
  \hline
  PFM+PCAL200+PCAH256-len45  & 200 &  69.9 & 57.7 & 42.3 & 56.6 \\
  PFM+PCAL200+PCAH256-len45  & 400 &  72.6 & \textbf{62.0} & 45.1 & 59.9 \\
  PFM+PCAL200+PCAH256-len45  & 600 &  72.6 & 54.9 & \textbf{54.9} & \textbf{60.8} \\
  PFM+PCAL150+PCAH256 & 600 &  \textbf{78.1} & 53.1 & 46.9 & 59.4 \\
 \hline
\end{tabular}
\end{table}
\subsection{Results}\label{subsec:tumRes}

The results shown in table Tab.~\ref{tab:resultsTUMA} correspond to \textit{experiment A}. 
We can see that all cases present high recognition rates ($\geq 99\%$), indicating the robustness of our method -- different values of the parameters only present maximum variations of $\pm 3\%$.

In the case of normal gait (\textit{N}) %
 we can see that lower values of $K$ are better due to the similarity of samples since both train and test sequences have the same walking conditions. Otherwise, in cases of carrying a bag (\textit{B}) and wearing coating shoes (\textit{S}), higher values of $K$ are necessary as sequences have different walking conditions and the algorithm requires larger dictionaries for representing the gait information.

The best case for normal gait (\textit{N}) is PFM+PCAL100+PCAH256 %
with $K=200$ where FM achieves an accuracy of $99.7\%$ (only one sequence is mismatched). For the remaining cases, the best configuration is the same but with a bigger dictionary ($K=600$) where FM reaches $99\%$ in both cases, \textit{B} and \textit{S}.

In table Tab.~\ref{tab:resultsTUMB} we can see the results of \textit{experiment B}. In this experiment, we obtain lower accuracy due to high variability between train and test sequences. Moreover, the low number of samples available for training the classifier makes harder to obtain discriminant information needed for the identification of subjects. To avoid this lack, we split the original training sequences into independent subsequences of $XX$ frames with $O$ frames of overlap. Thus, from one sequence we can obtain more subsamples that allow us to train a better classifier. Note that in Tab.~\ref{tab:resultsTUMB}, the rows follow the pattern PFM+PCAL200+PCAH256+lenXX where $XX$ corresponds to the frames of the partition. In our case, the best configuration is $XX = 45$ and $O = 10$ because partitions with a lower number of frames produces overfitting due to the high number of subsamples produced. On the other hand, partitions with higher number of frames produce only two subsamples with huge differences in the number of frames. %
If the row does not follow this pattern, it indicates that we have used the full sequence without partitions (e.g. bottom row of the table).

The best case for normal gait (\textit{TN}) is PFM+PCAL150+PCAH256 with $K=600$ and without time partition because train and test samples are similar. In this case, FM achieves an accuracy of $78.1\%$. For the case of carrying a bag (\textit{TB}) the best configuration is PFM+PCAL200+PCAH256+T45+O10 with $K=400$ where FM reaches an accuracy of $62.0\%$. Finally for wearing coating shoes case (\textit{TS}) the best result is $54.9\%$ with the configuration PFM+PCAL200+PCAH256+T45+O10 with $K=600$. As we can see in the results, time partition is useful for experiments where the variability of training and test samples is high. The same reasoning applies to the dictionary size as Tables~\ref{tab:resultsTUMA} and~\ref{tab:resultsTUMB} indicate: big dictionaries allow the algorithm to achieve better results in experiments over different conditions because a richer representation is obtained and consequently, a better generalization.

\begin{table}[t] %
\renewcommand{\arraystretch}{1.1}
\caption{\textbf{State-of-the-art on TUM GAID}. Percentage of correct recognition on TUM GAID for diverse methods. Each column corresponds to a different scenario. Best results are marked in bold. (See main text for further details).}
\label{tab:tumcomp}
\centering
\footnotesize
\setlength{\tabcolsep}{0.4em} %
\begin{tabular}{c|ccc|ccc|c}
\hline 
\textit{Method} & \textit{N} & \textit{B} & \textit{S} & \textit{TN} & \textit{TB} & \textit{TS} & \textit{Avg} \\ 
\hline
SDL~\cite{SDL} & - & - & - & 96.9 & - & - &  - \\ 
GEI~\cite{tumDB} & 99.4 & 27.1 & 52.6 & 44.0 & 6.0 & 9.0 & 56.0 \\
SEIM~\cite{SVIM} & 99.0 & 18.4 & 96.1 & 15.6 & 3.1 & 28.1 & 66.0 \\
GVI~\cite{SVIM} & 99.0 & 47.7 & 94.5 & 62.5 & 15.6 & 62.5 & 77.3 \\
SVIM~\cite{SVIM} & 98.4 & 64.2 & 91.6 & 65.6 & 31.3 & 50.0 & 81.4 \\
RSM~\cite{RSM} & \textbf{100.0} & 79.0 & 97.0 & 58.0 & 38.0 & \textbf{57.0} & 88.2 \\
\hline
PFM (this paper) & 99.7 & \textbf{99.0} & \textbf{99.0} & \textbf{78.1} & \textbf{62.0} & 54.9 & \textbf{96.0} \\
\hline 
\end{tabular}
\vspace{-0.04cm}
\end{table}
To put our results in context with other works, Tab.~\ref{tab:tumcomp} contains results from \textit{experiments A and B}. %
In particular, bottom row (PFM) shows our best results for each scenario, taken from Tabs~\ref{tab:resultsTUMA} and~\ref{tab:resultsTUMB}, %
along with the average performance. The first method (SDL) is specialized in temporal identification and the authors only report experiments for the case \textit{TN}, so we cannot obtain an average accuracy. Note how our approach improves on the state-of-the-art average from $88.2\%$ to $96.0\%$. The lowest accuracy is reached in \textit{TS} where we obtain a value only $2.1\%$ lower than best result. In the rest of cases (excluding specialized method in \textit{TN}) our method outperforms or obtains similar results to the state-of-the-art.

\section{Final Discussion}\label{sec:discussion}
We summarize here our main overall findings based on the experimental results obtained on the 
datasets. 

First of all, the results presented in  tables \ref{tab:resultsA},  
\ref{tab:resultsFeatDicsMOBO}, \ref{tab:casiaBcompMV} and \ref{tab:tumcomp} indicate 
that the proposed pipeline is a valid approach for gait recognition, obtaining a $100\%$ of correct 
recognition on the multiview setup on both AVAMVG and CMU MoBo datasets, and a $99.7\%$ on CASIA-B. 
In addition, the 
FV-based formulation surpasses the BOW-based one, as stated by other authors in the problem of 
image categorization~\cite{perronnin2010eccv}.
Moreover, the large dimensionality of the PFM can be drastically reduced by applying PCA, 
without worsening the final performance. 
For example, we can see in Tab.~\ref{tab:resultsA} that reducing the dimensions of the low-level 
motion descriptors to 100, and the final PFM to 256, allows to achieve a similar recognition rate 
but decreasing significantly the computational complexity ($\approx \times370$ smaller with 
$K=150$).

If we focus on the idea of spatially dividing the human body for computing different gait 
descriptors, 
the results in Tab.~\ref{tab:resultsA} show that the most discriminantive features are localized on 
the lower-body (row `PFM-H2'), what confirms our intuition (i.e. gait is mostly defined by the 
motion of the legs). In addition, although in a slight manner (see values in parenthesis), the 
upper-body features (row `PFM-H1') contribute to the definition of the gait as well.

Focusing on Tab.~\ref{tab:resultsTrX}, we can observe that PFM generalizes fairly well, 
as derived from the results obtained when testing on curved trajectories. 
Note that this kind of situations clearly benefits from the use of multiple cameras, as indicated 
by the low results yielded by single cameras and improved when combined.
Dealing with changing body viewpoints and deformations of the body parts highlights the importance 
of having a good person tracker able to properly group the person detections along time.
Actually, the results reported in this work on curved trajectories improve on the ones published in 
the conference version \cite{castro2014icpr}, thanks to the new stage that links broken 
tracks of persons (see Sec.~\ref{subsec:detection}).
With regard to the use of more than one level in PFM, we can see in Tab.~\ref{tab:resultsA} and 
Tab.~\ref{tab:resultsFeatDicsMOBO} that
similar results are obtained with the single- and two-level configurations. 
Although we tried an additional 
third level in the pyramid, the recognition rate did not increase. This fact indicates that, 
for most situations, is enough to use just a single vertical partition of the person's bounding-box 
to obtain very accurate results. However, for very complicated scenarios, as wearing coats on 
low resolution videos as CASIA-B, 
using two levels shows benefits as shown in Tab.~\ref{tab:casiaBmvnm}.

Concerning the contribution of the subtypes of descriptors in DCS, the experimental results 
suggest that (\textit{i}) not all of them are strictly necessary, (\textit{ii}) the 
\textit{normalized coordinates} can be safely omitted, and (\textit{iii}) in most cases the use of 
just the combination of \textit{div+curl} with \textit{curl+shear} is enough to achieve a very high 
recognition accuracy as shown in Tab.~\ref{tab:resultsFeatDics} and 
Tab.~\ref{tab:resultsFeatDicsMOBO}.

Although we have defined the gait recognition problem in a multiple-camera setup, the results 
reported in tables \ref{tab:resultsCam}, \ref{tab:resultsCamMOBO}, \ref{tab:casiaBmvnm} and \ref{tab:resultsTUMA} for the single camera case 
indicate that the proposed method is also valid for monocular environments. Thus, widening the 
range of application of our approach. 
A known limitation of our approach is the handle of 
trajectories perpendicular to the camera plane (i.e. perfectly frontal or backwards body 
viewpoint), where informative enough point trajectories cannot be computed. See for example, 
the case `cl' in Tab.~\ref{tab:casiaBmvnm}.
In our opinion, for such particular cases, the addition of shape-based features could help.

With regard to changes in appearance of people, we can say from the results on CASIA-B and TUM GAID, that our system is 
able to deal very well with people wearing bags, although improvement is needed with strong changes in clothing. 
In addition, although our system was not initially designed to deal with outdoor infrared images, the results of CASIA-C
clearly indicates that our PFM descriptor offers state-of-the-art results on that kind of data.

In summary, we can conclude that the proposed PFM allows to identify subjects 
by their gait by using as basis local motion (i.e. short-term trajectories) and coarse structural 
information (i.e. spatial divisions on the person bounding-box). Moreover, PFM does not need either 
segmenting or aligning the gait cycle of each subject as done in previous works.

\subsection{Speed of the proposed system}

To have an idea of the average speed of our system, we break down the time processing of the different 
stages comprising it. We have run this experiments on a state-of-the-art desktop computer with a 
CPU at $3.47$ GHz and $24$ GB of RAM. The non-parallel code is mostly written in Matlab with some pieces 
of code written in C++.   
The average time, in seconds, needed to process a video sequence of 50 frames from CASIA-B ($320 \times 240$ pixels) 
is as follows: a) dense tracks computation, 8.95; b) person detection, 54.2; c) person tracking plus
tracklets filtering, 0.62; d) PFM computation, 0.23; and, 
e) SVM classification, 0.02. Which makes a total of around $64$ seconds for that kind of 50-frames video sequence.
Clearly, the computational bottleneck is located on the person detection module. However, a GPU based
person detector%
\footnote{The HOG-based person detector available in OpenCV library can run at 60 pfs on our computers.} %
could be used instead. In the latter case, the system could achieve around $5$ fps. A future improvement could 
be the speed-up of the dense tracking module by restricting the computation of the tracklets to smaller
image regions guided by the person detector previously run, instead of processing the whole image frame
and, then, removing unuseful tracklets, as currently done in this work.

\section{Conclusions}\label{sec:conclusions}

We have presented a new approach for recognizing human gait in video sequences.
Our method builds a motion-based representation of the human gait by combining densely sampled local
features and Fisher vectors: the \textit{Pyramidal Fisher Motion}. 

The results show that PFM allows to obtain a high recognition rate on a multicamera setup on the 
evaluated datasets: AVAMVG, CMU MoBo and CASIA (sets B and C) and on a single camera setup like TUM GAID. In the case of AVAMVG, a perfect identification of the individuals 
is achieved when we combine information from different cameras and the subjects follow a straight 
path. In addition, our pipeline shows a good behaviour on unconstrained paths, as shown by 
the experimental results -- the model is trained on subjects performing straight walking 
trajectories and tested on curved trajectories.
In the case of CMU MoBo, we have seen as our method is able to deal with differences in speed, as well as with 
cases where the movement of the arms is not available (i.e. holding an object with both hands).
With regard to the PFM configuration, we have observed that it is beneficial to decorrelate 
(by using PCA) both the low-level motion features and the final PFM descriptor in 
order to achieve high recognition results and, in turn, decreasing the computational 
burden at test time -- the classification with a linear SVM is extremely fast on 256-dimensional vectors.
The experimental results also show that a single camera viewpoint is enough for recognition in
many cases, even using just a short time interval of the video sequence -- a PFM computed on around one second length
sequence allows
a perfect recognition on MoBo from a single viewpoint. 

Furthermore, the experiments on CASIA-B, CASIA-C and TUM GAID show that our system scales properly with the number of subjects, is able to handle changes in appearance and speed, as well as, is able to deal with recordings taken indoors, outdoors and during night.

Since we use a person detector to localize the subjects, the proposed system in not restricted 
to deal with scenarios with static backgrounds. Moreover, the motion features used in this paper 
can be easily adapted to non static cameras by removing the global affine motion as proposed 
by Jain et al. in~\cite{jain2013cvpr}.

In conclusion, PFM enables a new way of tackling the problem of gait recognition on single and multiple 
viewpoint scenarios, removing the need of using people segmentation as mostly done so far.

\section*{Acknowledgments}
This work has been partially funded by the Research Projects TIN2012-32952 and BROCA, both financed
by FEDER and the Spanish Ministry of Science and Technology; and by project TIC-1692 (Junta de Andaluc\'ia).
We also thank David L\'opez for his help with the setup of the AVAMVG dataset.
Portions of the research in this paper use the CASIA Gait Database collected by Institute of
Automation, Chinese Academy of Sciences.

\def\bibsection{\section*{References}}
\bibliographystyle{elsarticle-num}
\bibliography{longstrings,local,casiab,casiac,tum,pr,bibAVA}
\end{document}